%% file: main.tex
\documentclass{article}

\input{config.tex}

\usepackage[final]{corl_2025} % Uncomment for the camera-ready ``final'' version.

\title{SafeBimanual: Diffusion-based Trajectory Optimization for Safe Bimanual Manipulation}

\vspace{-5pt}
\author{
Haoyuan Deng$^{1}$, \quad
Wenkai Guo$^{1}$, \quad
Qianzhun Wang$^{1}$, \quad
Zhenyu Wu$^{2}$, \quad 
Ziwei Wang$^{1}$\thanks{Corresponding author.} \\
$^{1}$ Nanyang Technological University, $^{2}$ Beijing University of Posts and Telecommunications\\
\texttt{haoyuan.deng@ntu.edu.sg,
ziwei.wang@ntu.edu.sg}
}

\begin{document}
\maketitle

%===============================================================================
\input{sec/0_abstract}
\input{sec/1_instroduction}
\input{sec/2_related_work}
\input{sec/3_method}

\input{sec/4_experiments}

\input{sec/5_conclusion}
\input{sec/6_Limitaions}
%===============================================================================

% The acknowledgments are automatically included only in the final and preprint versions of the paper.
\acknowledgments{This work was supported by the MoE AcRF Tier 1 Seed Grant (RS17/24) and the NTU EEE Ignition Research Grant (024920-00001).}
% \clearpage

%===============================================================================

% no \bibliographystyle is required, since the corl style is automatically used.
\bibliography{reference}  % .bib
\appendix
\section*{Appendix}
\input{sec/appendix}

\end{document}

%% file: config.tex
\usepackage{graphicx}
\usepackage{amsmath}
\usepackage{amssymb}
\usepackage{multirow}
\usepackage[dvipsnames]{xcolor}
\usepackage{booktabs, tabularx, makecell, colortbl, array} 
\newcolumntype{Y}{>{\centering\arraybackslash}X}
\usepackage{siunitx} %
\usepackage{tcolorbox}  %
\usepackage{listings}
\usepackage{mathrsfs}
\usepackage{caption}                
\usepackage{float}                      
\usepackage{enumitem}
\usepackage{wrapfig}
\usepackage{soul}
\usepackage{upgreek}
\usepackage{vcell}
\usepackage{mwe} %
\usepackage{calc} %
\usepackage{pifont} %
\newcommand{\name}{SafeBimanual\xspace}
\usepackage{algorithm}
\usepackage{algpseudocode}
\usepackage{mathtools}
\usepackage{marvosym}   %
\newcommand{\ourcolor}{gray!9}
\definecolor{DarkGreen}{RGB}{0,128,0} % 更深的绿色

%% file: sec/0_abstract.tex
\begin{abstract}
    Bimanual manipulation has been widely applied in household services and manufacturing, which enables the complex task completion with coordination requirements.
    % While recent diffusion-based methods have demonstrated advantages in modeling multi-modal action distributions for bimanual manipulation, they overlook safety constraints of bimanual systems during policy inference, where highly dynamic coordination of arms raises the risk of collisions and object tearing. 
    Recent diffusion-based policy learning approaches have achieved promising performance in modeling action distributions for bimanual manipulation. However, they ignored the physical safety constraints of bimanual manipulation, which leads to the dangerous behaviors with damage to robots and objects.
    % To this end, we introduce \textbf{SafeBimanual}, a diffusion-based test-time bimanual trajectory optimization framework that embeds safety constraints into policy inference. 
    To this end, we propose a test-time trajectory optimization framework named \textbf{SafeBimanual} for any pre-trained diffusion-based bimanual manipulation policies, which imposes the safety constraints on bimanual actions to avoid dangerous robot behaviors with improved success rate.
    % Specifically, we design cost functions to constrain inconsistent bimanual coordination during the denoising process, which reduces bimanual behaviors conflict at each denoising step. Since different stages of bimanual coordination involve different safety constraints, we introduce a Safe Manager that leverages visual-language model (VLM) to flexibly schedules combination of cost functions, where keypoint-pairs selected by the VLM are used to compute differentiable costs. 
    Specifically, we design diverse cost functions for safety constraints in different dual-arm cooperation patterns including avoidance of tearing objects and collision between arms and objects, which optimizes the manipulator trajectories with guided sampling of diffusion denoising process. Moreover, we employ a vision-language model (VLM) to schedule the cost functions by specifying keypoints and pairwise relationship, so that the optimal safety constraint is dynamically generated in the entire bimanual manipulation process.
    %we employ the visual-language model (VLM) to generate keypoint pairs and design safety cost constraints between them to guide the trajectory generation, which enhances the feasibility of dual-arm action in each denoising step. Meanwhile, we propose a Safe-Manager to achieve flexible scheduling of safety costs and coordinate the constraints corresponding to different stages of bimanual manipulation, which avoids multiple constraint optimization conflicts.
    SafeBimanual demonstrates superiority on 8 simulated tasks in RoboTwin with a 13.7\% increase in success rate and a 18.8\% reduction in unsafe interactions over state-of-the-art diffusion-based methods. Extensive experiments on 4 real-world tasks further verify its practical value by improving the success rate by 32.5\%. \url{https://denghaoyuan123.github.io/SafeBimanip/}
    % Project website is at: \url{https://denghaoyuan123.github.io/Safebimanual./}
    
\end{abstract}

% Two or three meaningful keywords should be added here
\keywords{Bimanual Manipulation, Diffusion Policy, Trajectory Optimization} 

%% file: sec/1_instroduction.tex
%===============================================================================
\section{Introduction}
	
Bimanual manipulation \cite{grotz2024peract2benchmarkinglearningrobotic, zhang2024empoweringembodiedmanipulationbimanualmobile, lu2024anybimanual, zhao2023learning, taxonomy, grannen2023stabilizeactlearningcoordinate} is a fundamental capability for robots to perform real-world tasks in household service \cite{zhang2024empoweringembodiedmanipulationbimanualmobile}, surgical operations \cite{10149474} and manufacturing \cite{buhl2019dual}. Compared with the unimanual manipulation system, bimanual manipulators are able to handle more complex tasks such as cooking and sewing, where one arm stabilizes the target object and the other arm performs the demanded action. Therefore, learning generalizable and robust bimanual manipulation policy is desired in real-world robot manipulator deployment. However, the extremely high degrees of freedoms (DOFs) in bimanual manipulators poses challenges in policy learning due to the large policy space.
\input{pic/teaser}

To address this, hierarchical policy learning \cite{gupta2019relay, peng2022ase, triantafyllidis2023hybrid} has been widely studied where the original large space is decomposed with reduced complexity. To further decrease the complexity caused by action discretization, diffusion policy \cite{chi2023diffusion, yan2025m, ze20243d, ke20243d} models the action distribution via continuous denoising process for policy generation, which shows excellent scalability to high-dimensional output with the score-based learning. However, spatial and temporal coordination between arms are required in bimanual manipulation tasks, and the arm misalignment usually leads to dangerous behaviors including trajectory intersection of arms and inconsistent end-effector actions on the same objects because existing methods fail to consider diverse physical safety constraints in real-world deployment, which causes severe consequences such as collision and object tearing.

In this paper, we present a SafeBimanual method to impose the safety constraints of the real-world bimanual manipulation on diffusion policy to avoid dangerous robot behaviors. Different from existing diffusion policy which directly samples actions from denoising distribution \cite{chi2023diffusion, yan2025m, ze20243d, ke20243d}, our method acquires safe trajectory in test time by guided action sampling of pre-trained diffusion policies, which is efficiently adapted to different safety constraints across real-world deployment scenarios. More specifically, we design diverse cost functions for safety constraint in different bimanual cooperation patterns such as gripper movement consistency to avoid object tearing in symmetric cooperation and object collision avoidance in asymmetric cooperation. As illustrated in Figure~\ref{fig:teaser}, these cost functions are designed based on the dominant unsafe interaction patterns revealed by an unsafe bimanual manipulation taxonomy, which systematically analyzes 65 bimanual manipulation tasks across 7 benchmarks. The gradient of cost functions is applied to generate the guidance of diffusion denoising process. As the optimal cost functions should be dynamically generated in different stages of the entire bimanual manipulation process, we employ a vision-language model (VLM) to schedule the cost functions, where keypoints and corresponding pairwise relationship in the cooperation patterns are specified. Extensive experiments show that SafeBimanual increases the success rate in the RoboTwin simulator and the real world by 13.7\% and 32.5\% respectively compared with the state-of-the-art bimanual manipulation methods, as the unsafe interactions are reduced by 18.8\% and 30.0\% during the arm cooperation. Our contributions can be summarized as follows:
\begin{itemize}[leftmargin=*]
    \item We propose a test-time trajectory optimization framework for safe bimanual manipulation, which plays as a plug-and-play module for diffusion policy to efficiently adapt to diverse safety constraints across deployment scenarios.
    \item We present a VLM-based safety constraint scheduler by dynamically considering cooperation patterns, where keypoints and corresponding relationship are specified to generate optimal cost functions.
    \item We conducted extensive experiments in both simulators and real robots on a wide range of household tasks. The higher success rate and fewer unsafe interactions compared existing methods demonstrate the effectiveness of safety awareness in bimanual manipulation.
\end{itemize}

%offering enhanced stability, efficiency, and versatility across diverse domains such as household service \cite{zhang2024empoweringembodiedmanipulationbimanualmobile}, surgical operations \cite{10149474}, and manufacturing \cite{buhl2019dual}. Compared with the unimanual manipulation system, 

%===============================================================================

%% file: pic/teaser.tex
\begin{figure}[h]
  \centering
  \vspace{-20pt}
  \includegraphics[width=1.0\linewidth]{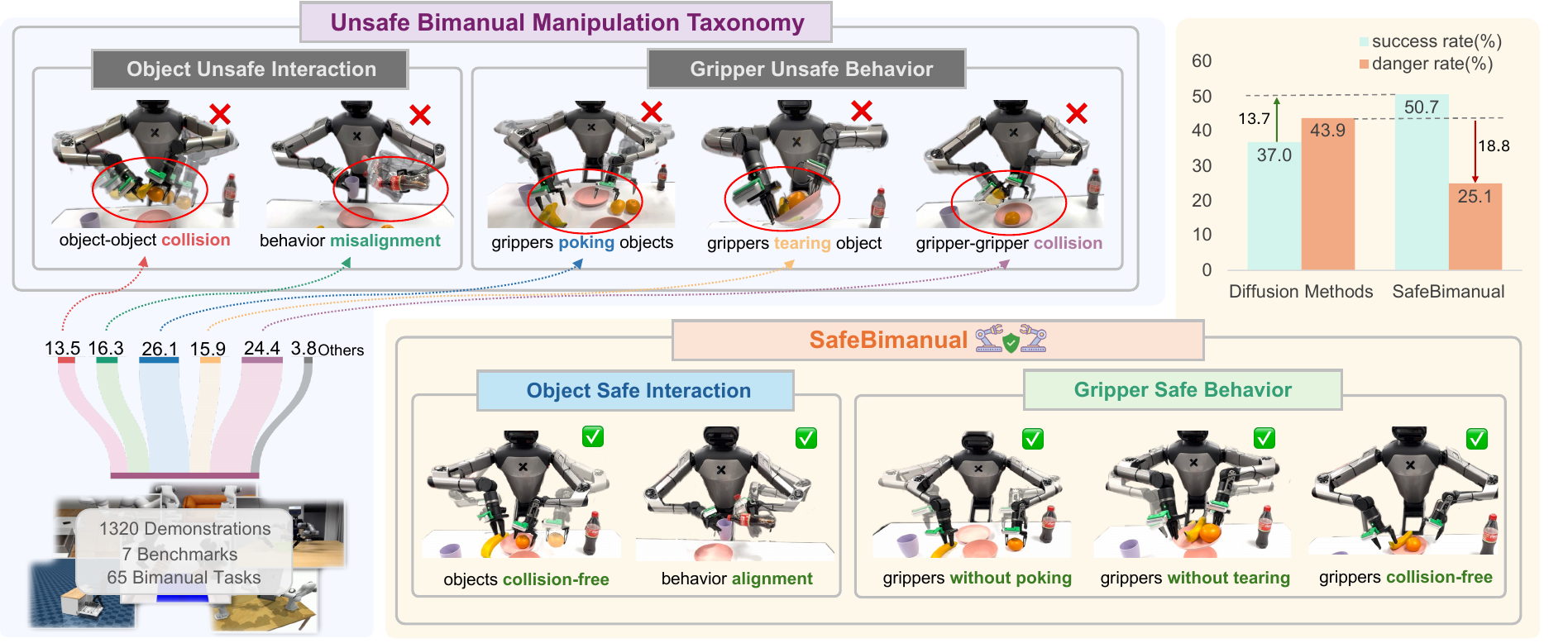}
  \vspace{-10pt}
  % \fbox{
  %       \begin{minipage}[c][5cm][c]{0.95\linewidth}
  %           \centering
  %           {\Large Dummy figure}
  %       \end{minipage}
  %   }
  \caption{\textbf{\name} imposes safety constraints into policy optimization to enable physically safe and task-effective bimanual manipulation across diverse scenarios.}
  \label{fig:teaser}
\end{figure}

%% file: sec/2_related_work.tex
\section{Related Work}
\label{sec:citations}

\textbf{Bimanual Manipulation.} Dual-arm systems \cite{smith2012dual, abbas2023systematic} can complete more complex manipulation tasks than single-arm manipulators \cite{lobbezoo2021reinforcement, iriondo2019pick} because of the higher flexibility and efficiency, which have been widely deployed in household service \cite{zhang2024empowering}, surgical operations \cite{wu2019coordinated} and industrial assembly \cite{li2019survey}. Early attempts applied reinforcement learning and imitation learning to generate bimanual manipulation policy with the guidance of designed reward and human expert demonstrations respectively. To deal with the large policy space caused by the high DOFs of dual-arm systems, the policy was learned hierarchically \cite{xie2020deep} where the bimanual tasks was decomposed into different single-arm manipulation tasks such as stabilizing and acting. However, learning the bimanual manipulation policy in the discrete action space still causes extremely high complexity.

\textbf{Diffusion Models in Robotics.} Inspired by the great success in image generation, diffusion models \cite{chi2023diffusion, ze20243d, ke20243d, xian2023chaineddiffuser} have aroused extensive interests in policy learning because of the expressive ability of multimodal action distributions and the scalibility to high-dimensional space, which improves the
stability, efficiency and success rate of bimanual manipulation policy learning. To leverage  richer information such as geometry, temporal relation and robot states in manipulation tasks, compact representations of 3D point cloud \cite{ze20243d, ke20243d}, videos \cite{xing2024survey} and proprioception \cite{wu2024tacdiffusion} are learned as conditions for diffusion policy generation. Meanwhile, diffusion sampling strategy is improved with higher efficiency to adapt to the limited computational resources in robots, where faster sampling paths \cite{lu2022dpm, lu2024manicm} or even one-step diffusion \cite{frans2024one} are explored for real-time applications. As spatial and temporal coordination between arms is required in bimanual manipulation, existing diffusion policy that ignores the safety constraints usually causes arm misalignment with dangerous behaviors such as collision and object tearing.

\textbf{Trajectory Optimization.} Trajectory optimization plays a pivotal role in robotic manipulation, enabling robots to generate feasible and efficient motions toward task goals \cite{kalakrishnan2011stomp, schulman2014motion}. Traditional methods typically impose geometric or contact constraints to ensure collision-free trajectories \cite{zucker2013chomp, manchester2019variational}, often rely on hand-crafted formulations for specific tasks and struggle to scale in high-DOF systems like bimanual manipulation. To address this, data-driven methods have emerged that learn task objectives and constraints directly from demonstrations or interaction data \cite{sundermeyer2021contact, urain2023se, chi2023diffusion}, improving adaptability in unstructured scenes. Benefiting from the rise of foundation models, recent works such as VoxPoser \cite{huang2023voxposer} and ReKeP \cite{huang2024rekep} demonstrate that leveraging VLM to infer affordances and constraints can further enhance generalization and planning efficiency \cite{huang2022language, driess2023palme, wu2025momanipvla}. In this work, SafeBimanual is inspired by these advances and tackles unsafe interactions in bimanual manipulation. We impose safety constraints into the diffusion denoising process via a VLM scheduler to avoid dangerous robot behaviors.

%===============================================================================

%% file: sec/3_method.tex
\section{Methodology}
\label{sec:method}

% In this paper, we present \textbf{SafeBimanual}, a plug-and-play test-time optimization module that imposes safety constraints into the deployment of diffusion-based policies for real-world bimanual manipulation. We introduce cost functions tailored to different cooperation patterns—such as synchronized motion in symmetric tasks and collision avoidance in asymmetric ones—to guide the denoising process and optimize safe manipulator trajectories. To adaptively schedule optimal cost functions across different manipulation stages, we develop a vision-language model (VLM)-based safety scheduler, which dynamically selects and compose cost functions based on detected relational keypoint pairs capturing critical spatial and functional relationships between dual-arm end-effectors and manipulated objects.

\begin{figure}[h]
  \centering
  \vspace{-10pt}
  \includegraphics[width=0.95\linewidth]{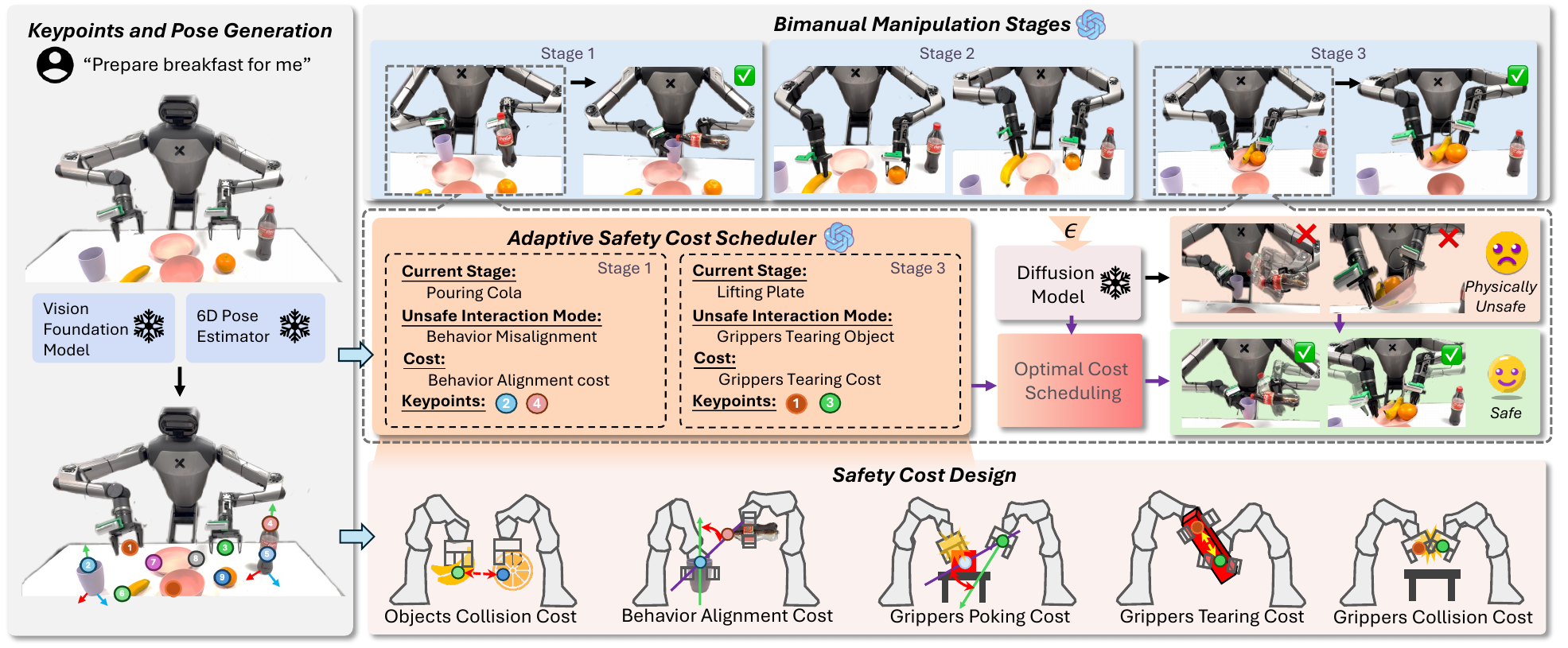}
  \vspace{-5pt}
  % \fbox{
  %       \begin{minipage}[c][5cm][c]{0.95\linewidth}
  %           \centering
  %           {\Large Dummy figure}
  %       \end{minipage}
  %   }
  \caption{\textbf{\name}. The framework integrates a Vision-Language Model (VLM)-based Adaptive Safety Cost Scheduler with stage-appropriate safety constraints. These constraints guide the diffusion denoising process to optimize dual-arm trajectories for safe and coordinated manipulation during deployment.}
  \label{fig:pipline}
\end{figure}

In this section, we first introduce problem statement (\ref{sec:Problem Formulation}). Then, we formulate imposing safety constraints on trajectories as guided sampling in denoising process of pre-trained diffusion policies (\ref{sec:Guided sampling}). 
To specify, we design cost functions based on the dominant unsafe interaction patterns revealed by an unsafe bimanual manipulation taxonomy to guide the denoising process (\ref{sec:Cost}). We further propose an adaptive scheduler to dynamically select optimal cost functions in different bimnaual task stages based on the specified cooperation patterns (\ref{sec:Schedul}). Our overall pipeline can be seen in Figure~\ref{fig:pipline}.

\subsection{Problem Statement}
\label{sec:Problem Formulation}

Bimanual manipulation involves coordinating two robotic arms to jointly manipulate objects, enabling the execution of complex tasks such as folding, assembly, or collaborative transport. 
Given the observation \(\mathcal{O}_t\) consisting of visual, proprioceptive, and optional instruction inputs at time \(t\), diffusion-based bimanual manipulation policies model the denoising distribution \(p(A_t \mid \mathcal{O}_t)\) to interactively predict optimal coordinated action sequences \(A_t = (a_{t+1}, \dots, a_{t+n})\), where each \(a_i \in \mathbb{R}^d = (q_i^\ell, q_i^r, g_i^\ell, g_i^r)\) encodes the joint positions \((q_i^\ell, q_i^r)\) and gripper states \((g_i^\ell, g_i^r)\) of the two arms.
However, directly sampling actions from the denoising distribution without considering physical safety constraints may result in coordination misalignment, inter-arm collisions, or object tearing during bimanual coordination, especially in real-world where precise spatial and temporal synchronization between arms is critical.
In light of this, we focus on imposing safety-aware guidance during the diffusion denoising process to promote physically feasible bimanual coordination. 
% We first decompose the task into sequential stages \(\{s_t\}\), and for each stage, we use a chain-of-thought (CoT) reasoning process to infer the corresponding unsafe interaction pattern based on the observation \(\mathcal{O}_t\). 
% xx 
We consider bimanual manipulation tasks decomposed into $K$ sequential stages $\{s_t\}_{t=1}^K$. 
For each stage $s_t$, the objective is to infer the potential unsafe interaction pattern 
and to quantify it using a set of differentiable safety costs $\mathcal{C}(A_t,\mathcal{P},s_t)$, 
where $\mathcal{P}$ denotes the relational keypoints between the two arms and the manipulated objects~\cite{huang2024rekep}, 
thereby steering the trajectory toward a physically safer version $A_t^{\text{safe}}$. 

% We use a chain-of-thought (CoT) reasoning process to decompose the task into $K$ sequential stages 
% $\{s_t\}_{t=1}^K$, and subsequently infer the unsafe interaction pattern for each stage $s_t$. 
% The Adaptive Safety Cost Scheduler then selects an optimal subset of the predefined differentiable cost terms 
% $\mathcal{C}_{\mathrm{sched}}(A_t, \mathcal{P}, s_t)$ to quantify unsafe behaviors, 
% where $\mathcal{P}$ denotes the relational keypoints between the two arms and the manipulated objects~\cite{huang2024rekep}.
% During this stage, at each timestep \(t\), we incorporate the gradients of the selected safety costs as guidance into the diffusion denoising process, progressively steering the trajectory toward a physically safer version \(A_t^{\text{safe}}\).

\subsection{Guided Sampling in Denoising Process.}
\label{sec:Guided sampling}
Following the Denoising Diffusion Probabilistic Model (DDPM) framework~\cite{ho2020denoising}, the policy generates actions by iteratively denoising from Gaussian noise:
\begin{equation}
p\bigl(A_t^{k-1} \mid A_t^k, O_t\bigr) = \mathcal{N}\!\left(A_t^{k-1};\, \mu(A_t^k, O_t, k),\, \Sigma\right),
\label{eq:reverse_kernel}
\end{equation}
where \(\mu(A_t^k, O_t, k)\) is the predicted mean, \(\Sigma\) is determined by the noise schedule, and \(k\) indexes the denoising timestep. To impose safety during generation, we extend classifier-guided diffusion to incorporate task-aware constraints through energy-based guidance. Since directly evaluating costs on noisy samples \(A_t^k\) is ineffective, we estimate the corresponding clean action chunk \(A_{0\mid k}\) using the noise prediction network \(\varepsilon_\theta\)~\cite{yu2023freedom}. We then compute a cost scheduled by the Adaptive Safety Cost Scheduler, denoted as \(\mathcal{C}_{\mathrm{sched}}(A_{0\mid k},\mathcal{P},s_t)\), and inject its gradient into the denoising update:
\begin{equation}
\label{eq:safety_guided_step}
A_t^{k-1} = \mu(A_t^k,O_t,k) - \rho_k\,\nabla_{A_k}\,\mathcal{C}_{\mathrm{sched}}(A_{0\mid k},\mathcal{P},s_t) + \sigma_k \varepsilon,\quad \varepsilon\sim\mathcal{N}(\mathbf{0},\mathbf{I}),
\end{equation}
where \(\rho_k\) controls the strength of safety guidance and \(\sigma_k\) is the noise schedule. This formulation steers trajectory \(A_t^{\mathrm{safe}}\) generation toward physically feasible and safe bimanual behaviors, guided by both task semantics and safety priors.

\subsection{Safety Cost Formulation}
\label{sec:Cost}
\textbf{Unsafe Bimanual Manipulation Taxonomy.} 
\label{Method:Taxonomy}
% To systematically characterise safe bimanual manipulation, we analyzed a comprehensive set of \textbf{65} bimanual manipulation tasks sourced from 7 benchmarks [xx]. As shown in Figure~\ref{fig:teaser}(c)[xx], Our analysis reveals that five recurring unsafe interaction patterns dominate the failure modes. 
To systematically characterize unsafe bimanual manipulation, we analysis 1,320 demonstrations from 65 tasks across 7 bimanual manipulation benchmarks \cite{james2020rlbench, mu2024robotwin, robomimic2021, chernyadev2024bigym, mittal2023orbit}, focusing exclusively on unsafe modes caused by dual-arm behaviors.
As illustrated in Figure~\ref{fig:teaser}, our study shows that over 96.2\% of such unsafe interactions fall into five representative categories, which we group into two physically grounded classes:
% We group these patterns into two primary classes: \textit{Objects Safe Interaction} and \textit{Gripper Safe Behavior}. Specifically, \textit{Objects Safe Interaction} encompasses scenarios involving object-object collisions and task-level objects behavior alignment. Meanwhile, \textit{Gripper Safe Behavior} addresses scenarios related to gripper interactions, including instances of gripper poking object, grippers tearing object, and collisions between dual grippers. 
% We group these patterns along the natural division between \emph{object–level} and \emph{gripper–level} interactions, which cleanly separates scene-centred collisions from end-effector behaviours and allows cost functions to be designed modularly for each side of the interaction. 
% Accordingly, we categorise safety issues into two classes:  
% \textit{Objects Safe Interaction}---covering object–object collisions and object–object behavior alignment; and  
% \textit{Gripper Safe Behavior}---covering gripper poking object, Gripper tearing object, and inter-gripper collisions. 
1) \textit{object unsafe interaction}: object-object collision and behavior misalignment;
2) \textit{gripper unsafe behavior}: gripper poking, tearing, and gripper-gripper collision.
% This taxonomy not only captures the vast majority of unsafe cases in our survey but also aligns with how corrective actions are deployed on the robot, enabling targeted and interpretable safety constraints.
This taxonomy emerges naturally from the physical characteristics of unsafe interactions: object-level unsafety typically involves misaligned object motion or unintended collisions, while gripper-level unsafety results from improper end-effector behaviors such as poking, tearing, or self-collision. The taxonomy forms a clear basis for defining modular safety constraints.

% \label{sec:Cost}
% \begin{wrapfigure}{c}{0.48\textwidth}
% \centering
% \includegraphics[width=0.92\linewidth]{example-image}
% % \vspace{-0.2em}
% \caption{\small \textbf{Bimanual Safe Cost.}}
% \vspace{-0.3em}
% \label{fig:visualization}
% \end{wrapfigure}

\textbf{Safety Cost.}
% Motivated by the above taxonomy, we design five differentiable cost terms, each targeting one of the dominant unsafe interaction patterns. Following the relational–keypoint paradigm~\cite{xx}, we represent the bimanual scene with a compact set of keypoints \(\mathcal{P}_t = \{k_t^\ell,\,k_t^r\}\), which indicate the most likely locations where unsafe interactions may occur. The estimated clean action chunk \(A_{0\mid k}\) comprises joint actions \(q^\ell\) and \(q^r\) for the left and right arms.
% % the keypoints are forward-kinematically projected to their future positions \(\widehat P(A_{0\mid k})\). 
% Each safety cost term \(c_i(A_{0\mid k}, \mathcal{P_t})\) is formulated as a differentiable function of pairwise keypoint relations.
Given keypoints \(\mathcal{P} = \{k_t^\ell,\,k_t^r\}\) identifying high-risk interaction regions of the left and right arm, we formulate five cost terms based on their spatial relationships, each addressing one of the five dominant unsafe interaction patterns.  
We express these constraints as differentiable functions of the estimated clean trajectory \(A_{0\mid k} = \bigl(q^\ell, q^r\bigr)\), so that gradients can be directly backpropagated into the action space. Following~\cite{manuelli2019kpam, huang2024rekep}, we assume rigidity between the end-effector and the grasped keypoints within an action chunk.  
The grasped keypoints include both the gripper tip and object-side points, which are computed as functions of the joint state \(q_t^i\) at each timestep:  
\begin{equation}
\label{costpre}
k_t^i = \mathcal{F}(q_t^i)\, T_i^{(0)}\, \hat{k}^i, \quad p^{\mathrm{tip}}_i = \mathcal{F}(q_t^i)\, T_{i}^{\mathrm{tip}}, \quad i \in \{\ell, r\}
\end{equation}
where \(\mathcal{F}(q_t^i)\) denotes the differentiable forward kinematics of arm \(i\),  
\(T_i^{(0)}\, \hat{k}^i\) is the static offset from the object keypoint to the end-effector, computed from the first action of action chunk,  
and \(T_i^{\mathrm{tip}}\) is the fixed transform from the gripper tip to the end-effector. Then each safety cost \(\mathcal{C}_{i}(A_{0\mid k}, \mathcal{P})\) is defined through differentiable metrics quantifying geometric relationships between keypoint:
% \paragraph{Object--object Collision.}

\textbf{1) Objects Collision Cost.}
\label{cost1}
To mitigate collision risks between manipulated objects, we impose a distance-based safety constraint \(\mathcal{C}_1\) over the object keypoints \(\mathcal{P}\) with the highest collision likelihood:
\begin{equation}
\mathcal{C}_1(A_{0\mid k}) = -\left\| k_t^\ell - k_t^r \right\|_2,
\end{equation}
where \(k_t^\ell, k_t^r\) are the transformed object keypoints of the left and right arms at timestep \(t\), computed in Eq (\ref{costpre}). By enforcing this constraint, we maintain a safe distance between the manipulated objects, effectively reducing the risk of unintended collisions or contact.
% \paragraph{Object--object collision (\(c_1\)).}To prevent undesired collisions between manipulated objects, we dynamically identify two keypoints \(k_t^\ell\) and \(k_t^r\) that are most likely to collide, and define a pairwise collision constraint across the predicted action chunk:
% \begin{equation}
% c_1(A_{0\mid k}) = -\left\| FK(q^\ell) T_{EEL}^{k_t^\ell} - FK(q^r) T_{EER}^{k_t^r} \right\|^2,
% \end{equation}
% where \(FK(\cdot)\) denotes the differentiable forward kinematics, and \(T_{EEL}^{k_t^l}\), \(T_{EER}^{k_t^r}\) are the rigid transformations from the end-effectors to the object keypoints \(k_t^l\) and \(k_t^r\), which are assumed remain constant throughout the action chunk.
% $\mathcal{C}_{1}$.

\textbf{2) Behavior Alignment Cost.}
% \paragraph{Object Behavior Misalignment.}
Take the example of a pouring water task where the left arm holds the bottle and the right arm holds the cup.
To enforce safe task-level alignment, which needs (1) to enforce axis alignment between the bottle keypoint \(k_t^\ell\) and the cup axis, and (2) to adjust the vertical offset concerning the cup position $k_t^r$.
The cost $\mathcal{C}_2$ can be denoted as:
\begin{equation}
\mathcal{C}_2(A_{0\mid k}) 
= \bigl\|(I - z\,z^\top)\, l_A\bigr\|^2 
\;+\;\lambda\,(z^\top l_A - h_0)^2, \quad l_A = k_t^\ell - k_t^r
\label{cost2}
\end{equation}
where \(z\in\mathbb{R}^3\) is the unit vector along the cup axis, and $h_0$ specifies the bottle-to-cup kepoints vertical offset.
Eq.~(\ref{cost2}) generalizes to bimanual tasks requiring spatial behavior alignment, such as insertion and stacking.
% \paragraph{Objects behavior misalignment (\(c_2\)).}  
% To ensure safe task‐level alignment—e.g. keeping the bottle mouth over the cup opening when pouring—we align the bottle keypoint \(k_t^\ell\) with the cup’s vertical axis and maintain a desired height above the cup pose center \(p_B = k_t^r\). Let $l_A = FK(q^\ell)\,T_{EEL}^{k_t^\ell} - FK(q^r)\,T_{EER}^{p_B}$. We then define:
% \begin{equation}
% c_2(A_{0\mid k}) 
% = \bigl\|(I - z\,z^\top)\, l_A\bigr\|^2 
% \;+\;\lambda\,(z^\top l_A - h_0)^2,
% \label{cost2}
% \end{equation}
% where \(z\in\mathbb{R}^3\) is the unit vector along the cup’s opening axis and \(h_0\) is the target mouth‐to‐rim height. In Eq.~(\ref{cost2}), we use the left hand holding the bottle and the right hand holding the cup as an example; this cost term generalizes to any task requiring behavior alignment (see Sec.~\ref{sec:Schedul}).

\textbf{3) Grippers Poking Cost.}
% \paragraph{Gripper Poking Object.}
\label{cost3}
To avoid unintended surface contact such as poking or scratching, we penalize deviations between the gripper’s approach direction and the keypoint–tip vector:
\begin{equation}
\mathcal{C}_3(A_{0\mid k}) = \bigl\|(I - aa^\top)\,(k_t^i - p^{\mathrm{tip}}_i)\bigr\|_2,
\end{equation}
where $a = Approach\bigl(\mathcal{F}(q^i_t)\bigr)$ is the gripper’s approach axis and \(p^{\mathrm{tip}}_i\) is the gripper tip position. $\mathcal{C}_3$ penalizes any component of the keypoint–tip vector that deviates from the intended approach axis.

\textbf{4) Grippers Tearing Cost.}
% \paragraph{Gripper Tearing Object.}
\label{cost4}
To prevent tearing a rigidly grasped object, we penalize deviations from the initial grasp width \(d_0\) between the two gripper tips:
\begin{equation}
\mathcal{C}_4(A_{0\mid k}) = \bigl(\|p^{\mathrm{tip}}_\ell - p^{\mathrm{tip}}_r\| - d_0\bigr)^2,
\end{equation}
This constraint helps prevent the unintended stretch or shear forces on the manipulated object.

% \paragraph{Gripper tearing object (\(c_4\)).}  
% When both grippers hold the same rigid object, their end-effector positions \(p_\ell\) and \(p_r\) should remain within a nominal distance \(d_0\) to avoid tearing.
% We penalize any deviation from the safe grasp width:
% \begin{equation}
% c_4(A_{0\mid k}) = \bigl(\|FK(q^\ell)\,T_{EEL}^{\mathrm{tip}} - FK(q^r)\,T_{EER}^{\mathrm{tip}}\| - d_0\bigr)^2,
% \label{cost4}
% \end{equation}
% where \(d_0\) is the initial distance between the two gripper tips when they first grasp the object.
\textbf{5) Grippers Collision Cost.}
% \paragraph{Gripper--Gripper Collision.}
\label{cost5}
To avoid end-effector collisions, we penalize proximity between the two gripper tips:
\begin{equation}
\mathcal{C}_5(A_{0\mid k}) = -\|p^{\mathrm{tip}}_\ell - p^{\mathrm{tip}}_r\|_2,
\end{equation}
This constraint maintains a minimum clearance between the grippers, reducing the risk of opposing forces between arms that may lead to mechanical deadlock or system failure.

% \paragraph{Gripper--gripper collision (\(c_5\)).}  
% To prevent the two end-effectors from colliding, we penalize their tips getting too close:
% \begin{equation}
% c_5(A_{0\mid k}) = -\|FK(q^\ell)\,T_{EEL}^{\mathrm{tip}} - FK(q^r)\,T_{EER}^{\mathrm{tip}}\|^2,
% \label{cost5}
% \end{equation}

\subsection{Adaptive Safety Cost Scheduler}
\label{sec:Schedul}
% Different phases of a bimanual task impose distinct safety priorities, such as collision avoidance during simultaneous moving of two objects and alignment during pouring. To schedule safety costs, we prompt VLM in a chain‐of‐thought (CoT) manner [xx], guiding the model through three reasoning steps: Identify unsafe pattern, Select optimal cost combination and Specify relational keypoints.
Different stages of a bimanual task impose distinct safety priorities, such as collision avoidance during simultaneous moving of two objects and alignment during pouring. 
To dynamically schedule safety constraints, we deploy a vision-language model (GPT-4o) in a structured chain-of-thought (CoT) process \cite{wei2022chain, zhang2024improve}, executing two sequential stages: 
% \textbf{Identify unsafe pattern} The VLM first considers the current observation \(\mathcal{O}_t\) with keypoints\(\mathcal{P}\), and task stage \(s_t\) to evaluate each predefined unsafe pattern \(p\in\mathcal{S}\), then picks \(\hat p_t = \arg\max_{p\in\mathcal{S}}\Pr(p\mid \mathcal{O}_t,\mathcal{P},s_t)\).  
\textbf{(1) Identify unsafe pattern.}  
Given the observation \(\mathcal{O}_t\), relational keypoints \(\mathcal{P}\), and current task stage \(s_t\), the VLM infers the most likely unsafe interaction pattern from a predefined taxonomy. This step enables the system to reason about safety-critical conditions grounded in real-world semantics.
\textbf{(2) Schedule cost terms with keypoints.}  
Conditioned on the predicted unsafe pattern, the VLM schedules a subset of relevant safety cost terms and identifies the corresponding relational keypoints. It activates only the necessary costs \(\mathcal{C}_i\) via a binary mask and assembles them into \(\mathcal{C}_{\mathrm{sched}}(A_{0\mid k}, \mathcal{P}, s_t)\).

For keypoints proposal, our relational keypoint set \(\mathcal{P}\) consists of: (i) semantically meaningful object keypoints in Cartesian space extracted using ReKeP~\cite{huang2024rekep}, tracked over time with CoTracker3~\cite{karaev24cotracker3} for temporal consistency;  
(ii) gripper tip positions \(p^{\mathrm{tip}}_\ell, p^{\mathrm{tip}}_r\) derived from forward kinematics; and  
(iii) object pose centers estimated via Omni6DPose~\cite{zhang2024omni6dpose}. Detailed are provided in the Appendix~\ref{app:Keypoints}.
 
For efficiency, we adopt a DDIM denoising strategy and apply safety guidance only in the final denoising steps (\(k \leq 3\)), during which the trajectory is refined using the gradient of the scheduled cost as shown in Eq.~\ref{eq:safety_guided_step}, progressively steering generation toward trajectories \(A_t^{\mathrm{safe}}\) that are both task-effective and physically safe.

%% file: sec/4_experiments.tex
\section{Experimental Results}
\label{sec:result}

We evaluate SafeBimanual through extensive experiments in both simulated and real-world settings. Our experiments addresses three key questions:  
\textbf{(1)} How effectively does SafeBimanual optimize diverse bimanual tasks while ensuring safety and task success (Section~\ref{sec:Simulation Results})?  
\textbf{(2)} Do the five safety cost terms sufficiently cover unsafe patterns, and can the scheduler adaptively select constraints across task stages (Section~\ref{sec:Ablations})?  
\textbf{(3)} What practical safety and generalization benefits does SafeBimanual bring in real-world applications (Section~\ref{sec:Real_world})?

\subsection{Experiment Setup}
\textbf{Simulation Benchmark.} 
We evaluate our framework on eight representative bimanual manipulation tasks from the RoboTwin~\cite{mu2024robotwin}, covering diverse dual-arm coordination patterns that span nearly all types of unsafe interactions.
See Appendix~\ref{appendix:Task_Description_sim} for detailed task descriptions.

\textbf{Baselines.} 
We apply SafeBimanual framework to optimize several representative diffusion-based policies, including 2D Diffusion Policy (DP), which predicts actions from images, and 3D Diffusion Policy (DP3), which uses compact 3D point cloud representations for imitation learning. We also include RDT-1b, a diffusion-based bimanual foundation model, as a strong baseline. All baselines are trained or fine-tuned with 50 expert demonstrations per task. See Appendix~\ref{app:params} for more details.
% Specifically, DP3 is trained for 3000 epochs with a batch size of 256, DP is trained for 300 epochs with a batch size of 128, and RDT-1b is fine-tuned for 10k steps on each task. 

\textbf{Evaluation Matrix.} \textbf{Success Rate (SR)} and \textbf{Danger Rate (DR)}. \textbf{SR} is the percentage of episodes (out of 100) completed successfully without any unsafe behavior. 
\textbf{DR} is the proportion of episodes where at least one unsafe interaction occurs, as defined by the taxonomy in Section~\ref{Method:Taxonomy}, which includes:
1) Object--Object Collision and 2) Gripper--Gripper Collision: detected via continuous collision monitoring in simulation.
% \textbf{Gripper poking object} triggered when a gripper tip makes undesired contact with non-grasped surfaces, as identified from visual observations.
3) Gripper Poking Object: triggered by unintended surface contact during grasping (e.g., poking or scratching).
4) Objects behavior misalignment: detected when the projected distance between two task-relevant keypoints on the plane orthogonal to the alignment axis exceeds the threshold \(d_{\text{align}} = 0.03\) m.
5) Gripper tearing object: flagged when the change in gripper tips distance for a rigidly held object exceeds \(d_{\text{tear}} = 0.04\) m.

\subsection{Simulation Results}
\label{sec:Simulation Results}
\textbf{Quantitative Experiments.} Experiment results are shown in Table~\ref{table:results}.  
(1) SafeBimanual consistently improves the performance of three state-of-the-art diffusion-based methods (DP, DP3, and RDT-1b) across eight simulated tasks in the RoboTwin environment. Specifically, it increases the success rate from 39.6\% to 50.7\% and reduces the danger rate from 43.9\% to 25.0\%, demonstrating the effectiveness of imposing safety constraints into the bimanual manipulation.
(2) SafeBimanual generalizes across diverse task types by handling different forms of unsafe interactions. Notably, it shows greater improvements in tasks involving multiple unsafe modes, such as Dual Shoes Place, Dual Bottles Pick (Hard), and Pour Water, where failures often result from unsafe interactions of gripper poking, object collisions, and behavioral misalignment.
In contrast, for tasks like Blocks Stack (Easy), where failures are primarily due to semantic mistakes (e.g., incorrect stacking order) rather than physical safety violations, SafeBimanual still reduces the danger rate (from 39.67\% to 23\%) but has limited impact on task success. This highlights that our method specifically addresses safety-critical failure modes. Overall, SafeBimanual sets a new performance benchmark in bimanual manipulation and demonstrates strong robustness across varying task complexities and unsafe interaction patterns.

\begin{figure}[t]
  \centering
  \vspace{-5pt}
  \includegraphics[width=1.0\linewidth]{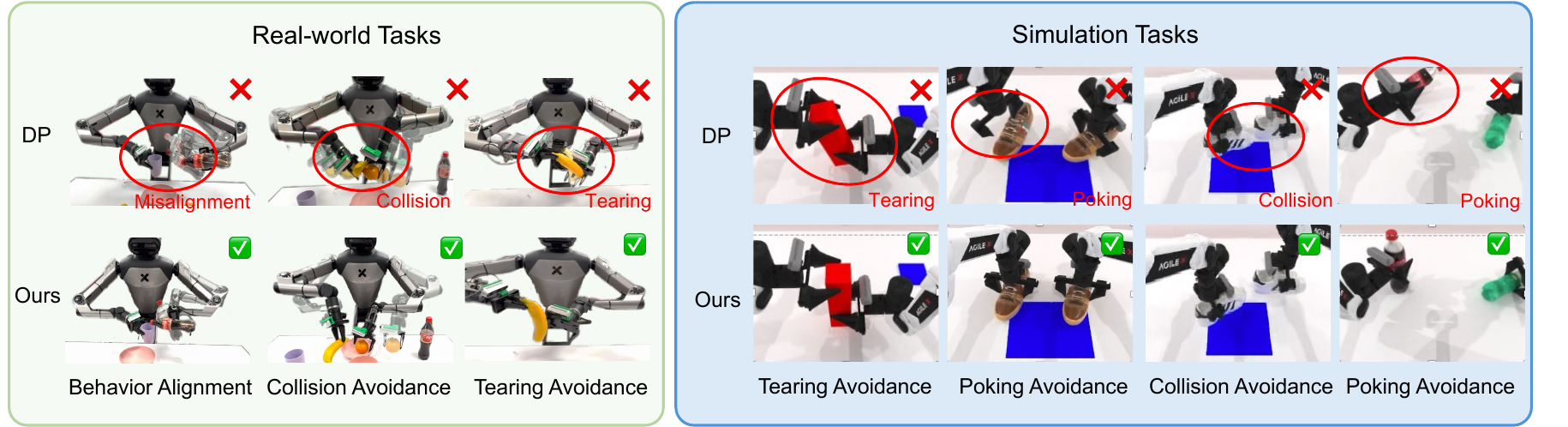}
  % \fbox{
  %       \begin{minipage}[c][4cm][c]{0.95\linewidth}
  %           \centering
  %           {\Large Dummy figure}
  %       \end{minipage}
  %   }
  \vspace{-15pt}
  \caption{\textbf{\name} integrates safety constraints to ensure robust and safe bimanual manipulation across diverse tasks.}
  \label{fig:visual}
  \vspace{-0.5cm}
  \end{figure}

\textbf{Qualitative Analysis.}  
As shown in Figure~\ref{fig:visual}, we visualize representative executions in both real-world and simulated environments.  
SafeBimanual successfully avoids diverse unsafe interaction patterns including collisions, misalignment, and gripper poking, while maintaining coordinated dual-arm behavior across different tasks.

\begin{table*}[t]
\centering
\footnotesize
\setlength{\tabcolsep}{5.2pt}
\renewcommand{\arraystretch}{1.1}
\caption{\textbf{Multi-Task Test Results in Simulator.}
SR \textcolor{green}{$\uparrow$} (Success Rate) and DR \textcolor{purple}{$\downarrow$} (Danger Rate) are evaluated over 100 episodes. SafeBimanual enables plug-and-play transfer for multiple diffusion-based policies, achieving a 13.8\% increase in success rate and a 18.9\% reduction in unsafe interactions.}
\vspace{-4pt}
\begin{tabularx}{\textwidth}{p{3.4cm}*{10}{Y}}
\toprule
\multirow{3}{*}{\textbf{Method}}
  & \multicolumn{2}{c}{\makecell{Dual Bottles\\Pick (Easy)}}
  & \multicolumn{2}{c}{\makecell{Dual Bottles\\Pick (Hard)}}
  & \multicolumn{2}{c}{\makecell{Block\\Handover}}
  & \multicolumn{2}{c}{\makecell{Blocks Stack\\(Easy)}}
  & \multicolumn{2}{c}{\makecell{Pour\\Water}}\\
\cmidrule(lr){2-3}\cmidrule(lr){4-5}\cmidrule(lr){6-7}\cmidrule(lr){8-9}\cmidrule(lr){10-11}
  & SR\textcolor{green}{$\uparrow$} & DR\textcolor{purple}{$\downarrow$}
  & SR\textcolor{green}{$\uparrow$} & DR\textcolor{purple}{$\downarrow$}
  & SR\textcolor{green}{$\uparrow$} & DR\textcolor{purple}{$\downarrow$}
  & SR\textcolor{green}{$\uparrow$} & DR\textcolor{purple}{$\downarrow$}
  & SR\textcolor{green}{$\uparrow$} & DR\textcolor{purple}{$\downarrow$}\\
\midrule
DP~\citep{chi2023diffusion} & 33\% & 23\% & 45\% & 38\% & 19\% & 13\% &  0\% & 40\% & 26\% & 30\%\\
\rowcolor{\ourcolor}\textbf{DP + \name}
            & \textbf{46\%} & \textbf{ 9\%} & \textbf{60\%} & \textbf{23\%}
            & \textbf{25\%} & \textbf{ 4\%} & \textbf{ 5\%} & \textbf{20\%}
            & \textbf{40\%} & \textbf{14\%}\\
\midrule
DP3~\citep{ze20243d} & 59\% & 19\% & 45\% & 37\% & 83\% & 17\% & 25\% & 46\% & 79\% & 21\%\\
\rowcolor{\ourcolor}\textbf{DP3 + \name}
            & \textbf{68\%} & \textbf{10\%} & \textbf{58\%} & \textbf{24\%}
            & \textbf{90\%} & \textbf{ 7\%} & \textbf{30\%} & \textbf{28\%}
            & \textbf{89\%} & \textbf{11\%}\\
\midrule
RDT-1b~\citep{liu2024rdt} & 56\% & 40\% & 39\% & 48\% & 94\% &  2\% & 14\% & 33\% & 41\% & 55\%\\
\rowcolor{\ourcolor}\textbf{RDT-1b + \name}
            & \textbf{66\%} & \textbf{30\%} & \textbf{47\%} & \textbf{39\%}
            & \textbf{96\%} & \textbf{ 0\%} & \textbf{21\%} & \textbf{21\%}
            & \textbf{59\%} & \textbf{41\%}\\
\bottomrule
\end{tabularx}

\vspace{5pt} 
\begin{tabularx}{\textwidth}{p{3.3cm}*{10}{Y}}
\toprule
\multirow{3}{*}{\textbf{Method}}
  & \multicolumn{2}{c}{\makecell{Pick Apple\\Handover}}
  & \multicolumn{2}{c}{\makecell{Dual Shoes\\Place}}
  & \multicolumn{2}{c}{\makecell{Diverse \\Bottles Pick}}
  % ▼ Average 跨 4 列
  & \multicolumn{4}{c}{\textbf{Average}}\\
\cmidrule(lr){2-3}\cmidrule(lr){4-5}\cmidrule(lr){6-7}\cmidrule(lr){8-11}
  % ▼ 第二行表头：SR/DR 各跨 2 列
  & SR\textcolor{green}{$\uparrow$} & DR\textcolor{purple}{$\downarrow$}
  & SR\textcolor{green}{$\uparrow$} & DR\textcolor{purple}{$\downarrow$}
  & SR\textcolor{green}{$\uparrow$} & DR\textcolor{purple}{$\downarrow$}
  & \multicolumn{2}{c}{SR\textcolor{green}{$\uparrow$}}
  & \multicolumn{2}{c}{DR\textcolor{purple}{$\downarrow$}}\\
% \cmidrule(lr){8-9}\cmidrule(lr){10-11}
%   % ▼ 第三行表头：把 SR 的两列命名为“值 / Δ”，DR 同理
%   &  & 
%   &  & 
%   &  & 
%   & \makecell{Value} & \makecell{$\Delta$}
%   & \makecell{Value} & \makecell{$\Delta$}\\
\midrule

% ▼ 数据行：最后 4 列依次为 SR值 / SR增量 / DR值 / DR增量
DP~\citep{chi2023diffusion}
  & 31\% & 63\%
  &  5\% & 88\%
  &  7\% & 70\%
  & \multicolumn{2}{c}{20.7\%}
  & \multicolumn{2}{c}{45.6\%} \\

\rowcolor{\ourcolor}\textbf{DP + \name}
  & \textbf{57\%} & \textbf{23\%}
  & \textbf{31\%} & \textbf{34\%}
  & \textbf{24\%} & \textbf{40\%}
  & \multicolumn{2}{c}{\textbf{36.0\%}\scriptsize (\textcolor{green}{ $\uparrow$ 15.3})}
  & \multicolumn{2}{c}{\textbf{20.9\%}\scriptsize (\textcolor{purple}{ $\downarrow$ 24.7})}\\

\midrule
DP3~\citep{ze20243d}
  & 82\% & 18\%
  & 24\% & 70\%
  & 34\% & 54\%
  & \multicolumn{2}{c}{53.9\%}
  & \multicolumn{2}{c}{35.3\%} \\

\rowcolor{\ourcolor}\textbf{DP3 + \name}
  & \textbf{94\%} & \textbf{ 3\%}
  & \textbf{53\%} & \textbf{38\%}
  & \textbf{56\%} & \textbf{32\%}
  & \multicolumn{2}{c}{\textbf{67.3\%}\scriptsize (\textcolor{green}{ $\uparrow$ 13.4})}
  & \multicolumn{2}{c}{\textbf{19.1\%}\scriptsize (\textcolor{purple}{ $\downarrow$ 16.2})}\\

\midrule
RDT-1b~\citep{liu2024rdt}
  & 33\% & 57\%
  &  7\% & 90\%
  &  7\% & 81\%
  & \multicolumn{2}{c}{36.4\%}
  & \multicolumn{2}{c}{50.8\%} \\

\rowcolor{\ourcolor}\textbf{RDT-1b + \name}
  & \textbf{45\%} & \textbf{41\%}
  & \textbf{26\%} & \textbf{56\%}
  & \textbf{31\%} & \textbf{54\%}
  & \multicolumn{2}{c}{\textbf{48.9\%}\scriptsize (\textcolor{green}{ $\uparrow$ 12.5})}
  & \multicolumn{2}{c}{\textbf{35.3\%}\scriptsize (\textcolor{purple}{ $\downarrow$ 15.5})}\\

\bottomrule
\end{tabularx}

\vspace{-0.2cm}
\label{table:results}
\end{table*}

\subsection{Ablations}
\label{sec:Ablations}
\name integrates two key components: a set of safety cost functions and an adaptive scheduling mechanism for selecting appropriate constraints during execution. We conduct ablations by applying SafeBimanual to optimize diffusion policy~\cite{chi2023diffusion}.

\textbf{Safety Cost Ablation.} We perform leave-one-out ablations over the five safety cost terms and report results in Table~\ref{tab:ablation_combined}. Removing any single cost consistently increases the overall Danger Rate, confirming the importance of each constraint. The object--object collision cost has the most significant impact, as such collisions are common across all tasks. In contrast, the gripper tearing cost has a smaller effect, since tearing mainly arises in specific tasks like Block Handover and Pick Apple Handover.

\begin{wraptable}{r}{0.48\textwidth}  % 'r' for right, adjust width as needed
% \vspace{-1em}
\begin{minipage}{\linewidth}
\centering
\footnotesize
\setlength{\tabcolsep}{4pt}
\renewcommand{\arraystretch}{1.0}
\caption{\textbf{Ablation.} Each ablation setting is tested over 20 episodes across 8 tasks.}
% \vspace{1.2pt}
\begin{tabular}{lcc}
\toprule
\multicolumn{1}{l}{\textbf{Method}} &
\multicolumn{2}{c}{\textbf{Average}} \\[-0.2em]
\cmidrule(lr){2-3}
& SR\,(\%)\textcolor{green}{$\uparrow$} & DR\,(\%)\textcolor{purple}{$\downarrow$} \\\midrule
w/o $\mathcal{C}_1$ & 31.8 & 24.4\\
w/o $\mathcal{C}_2$ & 33.8 & 23.1\\
w/o $\mathcal{C}_3$ & 25.0 & 37.5\\
w/o $\mathcal{C}_4$ & 33.1 & 21.3\\
w/o $\mathcal{C}_5$ & 30.0 & 24.4\\
\midrule
w/o VLM (fixed weights)   & 19.4   & 42.5\\
w/ VLM w/o CoT            & 25.6   & 36.9\\
\midrule
\rowcolor{\ourcolor}\textbf{SafeBimanual} & \textbf{35.5} & \textbf{19.8}\\
\bottomrule
\end{tabular}
\label{tab:ablation_combined}
\end{minipage}
\vspace{-1em}
\end{wraptable}

\textbf{Cost Scheduling Ablation.} 
We compare three scheduling strategies: (1) a fixed-weight baseline that activates all costs equally, (2) a VLM-based scheduler without chain-of-thought (CoT) reasoning, and (3) our proposed CoT-VLM scheduler that dynamically selects cost terms based on inferred unsafe patterns.  
As shown in Table~\ref{tab:ablation_combined}, fixed-weight scheduling often fails to adapt to task-specific safety needs and may introduce conflicting constraints, which can hinder task execution and reduce success rates. In contrast, our adaptive CoT scheduler consistently improves success rate and reduces Danger Rate by selecting stage-appropriate cost combinations.

% These results validate our core motivation: effectively preventing unsafe interactions in bimanual manipulation requires not only a diverse set of task-relevant safety constraints but also dynamic, context-aware scheduling. Static or naive cost selection strategies are insufficient in real-world scenarios where dominant unsafe patterns shift throughout the manipulation process. Our design, integrating fine-grained cost functions with CoT-driven adaptive scheduling, enables SafeBimanual to robustly maintain physical safety while maximizing task success.
\subsection{Real-World Experiments}
\label{sec:Real_world}
% \begin{wraptable}{r}{0.6\textwidth}
% \vspace{-1em}
% \begin{minipage}[t]{\linewidth}
% \centering
% \footnotesize
% \setlength{\tabcolsep}{2.5pt}
% \renewcommand{\arraystretch}{1.1}
% \captionof{table}{\textbf{Real-World Results.}}
% \label{tab:vla_reasoner_results}
% \begin{tabular}{lcccccccc}
% \toprule
% \multicolumn{1}{l}{\textbf{Method}} &
% \multicolumn{2}{c}{\makecell{\textbf{Pass}\\\textbf{Banana}}} &
% \multicolumn{2}{c}{\makecell{\textbf{Pour}\\\textbf{Water}}} &
% \multicolumn{2}{c}{\makecell{\textbf{Wipe}\\\textbf{Bowl}}} &
% \multicolumn{2}{c}{\makecell{\textbf{Place}\\\textbf{Fruits}}} \\
% \cmidrule(lr){2-3}\cmidrule(lr){4-5}\cmidrule(lr){6-7}\cmidrule(lr){8-9}
% & SR & DR & SR & DR & SR & DR & SR & DR \\
% \midrule
% DP~\cite{chi2023diffusion} & 70\% & 30\% & 40\% & 60\% & 30\% & 30\% & 20\% & 60\% \\
% \rowcolor{\ourcolor}
% \textbf{+SafeBimanual} & \textbf{90\%} & \textbf{0\%} & \textbf{70\%} & \textbf{30\%} & \textbf{70\%} & \textbf{10\%} & \textbf{60\%} & \textbf{20\%} \\
% \bottomrule
% \end{tabular}
% \vspace{0.3em}
% \end{minipage}
% \vspace{-1em}
% \end{wraptable}
\begin{wrapfigure}{r}{0.6\textwidth}  % r 表示靠右, l 表示靠左
  \centering
  % \vspace{-10pt}
  \includegraphics[width=0.6\textwidth]{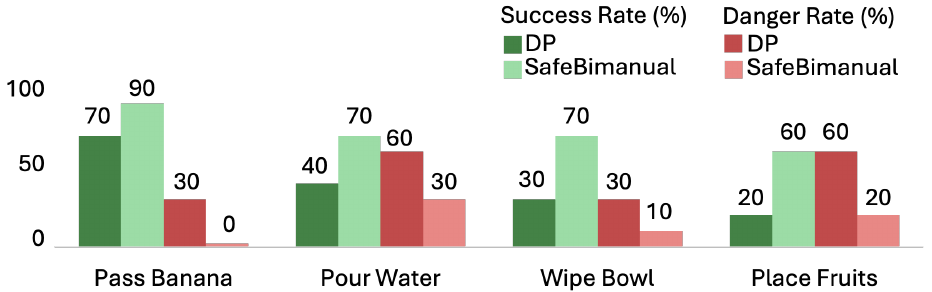}
  % \vspace{-1.6em}
  \caption{\textbf{\name} Real-World Results.}
  \captionsetup{aboveskip=2pt, belowskip=0pt}
  \label{fig:real_world}
\end{wrapfigure}
We evaluate SafeBimanual on the Galaxea-R1 humanoid robot using four real-world bimanual tasks that span all unsafe interaction types in our taxonomy. 
SafeBimanual is deployed to optimize pre-trained diffusion policies, resulting in higher success rates and fewer unsafe interactions compared to the original policies as shown in Figure~\ref{fig:real_world}. To further assess long-horizon performance, we introduce a challenging \emph{Prepare-Breakfast} benchmark that combines all five unsafe interaction modes. Fine-tuning iDP3~\cite{ze2024humanoid_manipulation} with SafeBimanual enhances object-level generalization and reduces compounding errors over extended sequences. Detailed task settings are provided in the Appendix~\ref{appendix:Task_Description_real}.

% \textbf{Results and Analysis.} As shown in Table~\ref{tab:vla_reasoner_results}, SafeBimanual achieves higher success rates and fewer unsafe interactions compared to the diffusion policy, which further verify its practical value.

% \textbf{Long-Horizon Task.} 
% We further devise a challenging long-horizon benchmark, \emph{Prepare-Breakfast}, which chains together all five unsafe interactions. Using SafeBimanual to fine-tune the iDP3 \cite{ze2024humanoid_manipulation} policy shows that SafeBimanual improves object-level generalization while curbing long-horizon compounding error.

%% file: sec/5_conclusion.tex
\section{Conclusion}
\label{sec:conclusion}

In this work, we present SafeBimanual, a test-time trajectory optimization framework that enhances the safety of diffusion-based bimanual manipulation policies to enable safe interactions. We design safety constraints to guide the trajectory denoising process and prevent unsafe behaviors. In addition, we leverage a vision-language model (VLM) to dynamically schedule optimal constraint combinations based on keypoints and their relational structures, avoiding unsafe interactions such as collisions, tearing, and misalignment. Our approach generalizes across diverse dual-arm cooperation patterns and achieves consistent performance improvements in both simulated and real-world tasks. These results highlight the practical value of incorporating physical safety reasoning into policy optimization for reliable robotic manipulation in complex environments.

%===============================================================================

%% file: sec/6_Limitaions.tex
\newpage
\section{Limitations}
While SafeBimanual significantly reduces unsafe interaction rates and improves task success across a wide range of scenarios, it does not guarantee complete safety under all conditions. In extreme or highly cluttered environments, residual unsafe behaviors may still occur due to the limitations of policy expressiveness or constraint coverage.
Moreover, SafeBimanual is specifically designed to mitigate unsafe physical interactions (e.g., collisions, poking, tearing) and is less effective in addressing semantic task failures, such as incorrect execution order—for instance, placing blocks in the wrong color sequence.
Our method also relies on visual keypoints and estimated poses, which may suffer from occlusion or noise. To mitigate this, we employ 2D keypoint tracking with a fixed grasp-offset assumption to improve robustness once an object is held; implementation details are provided in the Figure~\ref{fig:keypoint_proposal}.
Additionally, the quality of constraint scheduling depends on the reasoning ability of the vision-language model (VLM). While the CoT-based scheduler shows strong performance overall, it occasionally generates incomplete or suboptimal constraints due to limitations in VLM understanding. We expect this aspect to improve with continued progress in large-scale pre-trained models.
Lastly, we observe that in subjective tasks such as object folding or table setting, the VLM may miss fine-grained steps aligned with human preferences, which are inherently ambiguous and task-dependent.

%% file: sec/appendix.tex
% \documentclass{article}
% \usepackage{amssymb}
% \usepackage{booktabs, tabularx, array}   % nice tables
% \usepackage{caption}                    % \captionof + \captionsetup
% \usepackage{float}                      % defines the [H] float specifier
% \usepackage{tabularx}
% \input{sec/config2}
% % \usepackage{corl_2025} % Use this for the initial submission.
% % \usepackage[final]{corl_2025} % Uncomment for the camera-ready ``final'' version.
% \usepackage[preprint]{corl_2025} % Uncomment for pre-prints (e.g., arxiv); This is like ``final'', but will remove the CORL footnote.

% \title{SafeBimanual Supplementary}

% % The \author macro works with any number of authors. There are two
% % commands used to separate the names and addresses of multiple
% % authors: \And and \AND.
% %
% % Using \And between authors leaves it to LaTeX to determine where to
% % break the lines. Using \AND forces a line break at that point. So,
% % if LaTeX puts 3 of 4 authors names on the first line, and the last
% % on the second line, try using \AND instead of \And before the third
% % author name.

% % NOTE: authors will be visible only in the camera-ready and preprint versions (i.e., when using the option 'final' or 'preprint'). 
% % 	For the initial submission the authors will be anonymized.
% \begin{document}
% \maketitle
%===============================================================================
\appendix

%===============================================================================

\section{Video Demo}
A short video shows \textbf{SafeBimanual} imposing safety constraints to a diffusion-based policy for bimanual manipulation. The clip explains the idea, shows each step of the method, and ends with the long‑horizon \emph{Prepare Breakfast} task completed safely. Enjoy!

\section{Task Description}
\label{appendix:Task_Description}
\begin{figure}[h]
  \centering
  \vspace{-10pt}
  \includegraphics[width=1.0\linewidth]{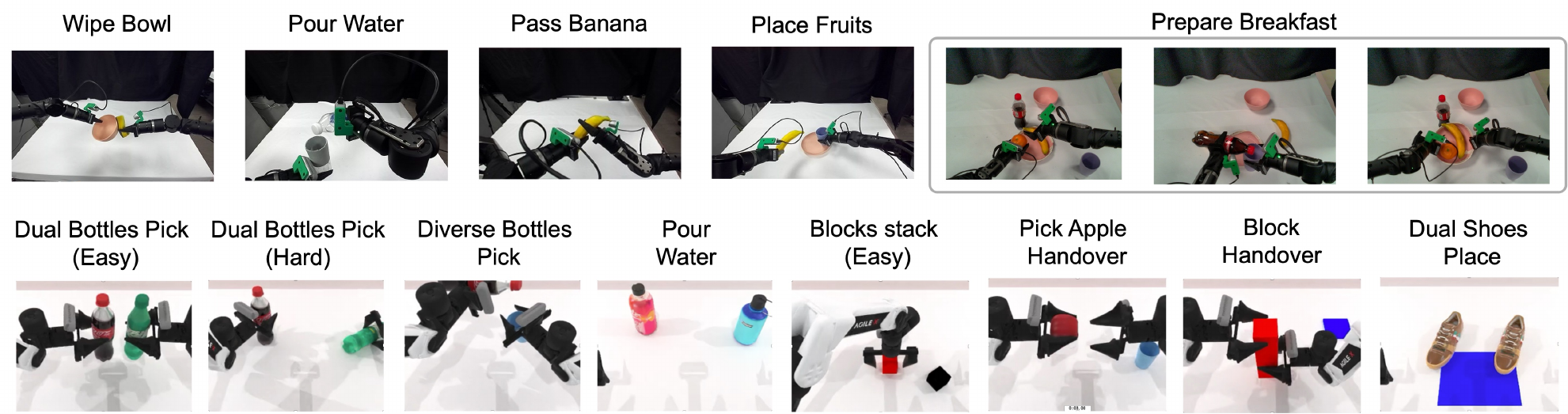}
  \vspace{-10pt}
  % \fbox{
  %       \begin{minipage}[c][5cm][c]{0.95\linewidth}
  %           \centering
  %           {\Large Dummy figure}
  %       \end{minipage}
  %   }
  \caption{Snapshot of all simulation(second line) and real-world tasks(first line).}
  \label{fig:task}
\end{figure}

\subsection{Simulation Tasks}
\label{appendix:Task_Description_sim}

\textbf{Dual Bottles Pick (Easy):} Two upright bottles are randomly placed on the left and right sides of the table. Each arm must grasp its assigned bottle, lift it, and transport both to a shared target area. Success means both bottles end up upright and stably positioned within the target zone without dropping.

\textbf{Dual Bottles Pick (Hard):} Two bottles are randomly oriented (upright, tilted, or lying down) and placed left and right. Each arm must identify, grasp, and lift its bottle despite the varied poses, then place both upright in the target area. Success requires both bottles to be upright and stably located in the goal region.

\textbf{Block Handover:} A long rectangular block lies on the left side of the table. The left arm must pick it up, transfer it to the right arm, and the right arm must place it at a marked goal on the right. Success is achieved when the block is handed over and placed stably at the goal.

\textbf{Blocks Stack (Easy):} One red and one black cube are scattered randomly. The robot must stack the black cube on top of the red cube in the correct order. Success is defined by a stable vertical stack in the specified color sequence.

\textbf{Pour Water:} A filled bottle sits on the left and an empty cup on the right. The left arm must grasp and tilt the bottle over the cup held by the right arm to pour water. Success requires water to transfer into the cup without spilling outside.

\textbf{Pick Apple Handover:} An apple is placed at a random location on the left side of the table. The left arm must grasp the apple, transfer it to the right arm, and fully open its gripper to complete the handover. Success is achieved when the apple is securely held by the right arm and the left gripper is fully open with no residual contact.

\textbf{Dual Shoes Place:} Two shoes from the same pair, each with a distinct design, are placed randomly on the left and right sides of the table. The robot’s left and right arms must each grasp their assigned shoe and deposit it into the blue target area, ensuring the toe of each shoe points toward the table’s left or right edge.  

\textbf{Diverse Bottles Pick:} Two bottles of different shapes and sizes are randomly positioned. Each arm must pick its assigned bottle and deliver both to a common target region. Success means both bottles are placed upright and stably in the goal area.

\subsection{Real-World Tasks}
\label{appendix:Task_Description_real}
\textbf{Pass Banana:} A banana lies on the left side of the workspace. The left arm must grasp the banana, pass it to the right arm, which places it into the other side of table. Success is achieved when the banana is transferred intact and stably positioned within the target region.

\textbf{Pour Water (Real):} A filled bottle and an empty cup are arranged on a table. The robot must coordinate both arms to pour water from the bottle into the cup. Success requires sufficient water transfer with minimal spillage.

\textbf{Wipe Bowl:} A bowl sits on the left and a sponge on the right. The left arm must lift and stabilize the bowl while the right arm wipes its interior with the sponge. Success is defined by a clean bowl and no drops or excessive force.

\textbf{Place Fruits:}  
A juice cup is positioned to the right of a plate and a banana to the left. The right arm must first grasp and place the cup onto the plate, and then the left arm must grasp and place the banana beside it. Success is achieved when both the cup and banana rest stably in their intended positions on the plate.

\textbf{Prepare Breakfast:} A cola bottle, an empty cup, an orange, and a banana are arranged. In Phase 1 the robot pours water from the bottle into the cup; in Phase 2 it places the orange and banana onto a plate; in Phase 3 it lifts and carries the loaded plate to a target area. Success requires each phase to complete without spillage, misplacement, or dropping.

\section{Pseudo-code for \name} 
% \textbf{Stage Decomposition and Constraint Generation.}  
% We first decompose each bimanual task into discrete stages \(\{s_t\}\) (e.g., pour water, place the fruits, lift plate). For simple tasks, we generate all stage‐specific safety constraints and corresponding termination check functions at the outset, using these termination checks to switch on the next stage’s cost guidance. For long‐horizon tasks, we instead re‐invoke the VLM at each stage termination to produce fresh safety constraints, which adds latency but improves robustness.

% \textbf{Pseudo‐Code Explanation.}  
% Algorithm~\ref{alg:safebimanual_ddim} implements our test‐time trajectory optimization: at each stage \(s_t\), we infer the unsafe interaction pattern via CoT reasoning, schedule the relevant safety costs \(\mathcal{C}_i\), and then perform DDIM denoising from step \(K\) down to \(1\). Crucially, we inject the gradient of the scheduled cost only in the final 30\% of steps (\(k \le 0.3K\)); earlier steps use pure denoising updates without safety guidance.

% \textbf{Why only the last 30\% of steps?}  
% Guidance added early in the denoising process is overwhelmed by high noise levels and can reduce trajectory diversity, since the policy initially explores a broad distribution. In later steps, when the sample has converged toward a narrow, near‐deterministic manifold, gradient‐based corrections more effectively steer the trajectory toward physically safe behaviors without being drowned out by stochastic noise.

% \paragraph{Algorithm Workflow.} 
\textbf{Algorithm Workflow.}  
At runtime (Algorithm~\ref{alg:safebimanual_ddim}), SafeBimanual first decomposes the task into discrete stages \(\{s_t\}\), each governed by a termination condition. For simple tasks, we pre-select all stage-specific safety constraints and their associated termination checks, automatically switching to the next stage’s cost guidance when its condition is met. For long-horizon tasks, we instead re-invoke the VLM at each stage boundary to generate fresh safety primitives—adding delay but enhancing adaptability to dynamic scenes.

Once the active stage \(s_t\) is determined, SafeBimanual applies CoT-VLM reasoning to the current observation \(\mathcal{O}_t\) and keypoints \(\mathcal{P}\) to identify the prevailing unsafe interaction pattern. The Adaptive Safety Cost Scheduler uses this result to produce a binary mask \(\alpha\) that activates only the relevant cost terms \(\{\mathcal{C}_i\}\). We then sample initial noise \(A_t^K\) and run DDIM denoising from \(k=K\) down to \(1\). To balance exploration and safety, we inject cost-gradient guidance only in the final \(M\) steps (\(k \le M = 0.3K\)); early steps remain pure denoising to preserve trajectory diversity, while later guidance steers the trajectory toward the safe output \(A_t^{\mathrm{safe}}\) once noise levels are sufficiently low.  

\label{appendix:inference}
\begin{algorithm}[t]
\caption{Safety‐Guided Sampling Process for Bimanual Manipulation}
\label{alg:safebimanual_ddim}
\begin{algorithmic}[1]
\Require Pretrained diffusion-based policy $\epsilon_\theta$~\cite{yu2023freedom}, safety costs $\{\mathcal{C}_i\}$, taxonomy $\mathcal{S}$, keypoints $\mathcal{P}$, scheduler VLM~\cite{wei2022chain,zhang2024improve}, total denoising steps $K$, guided denoising steps $M$, execution horizon $m$
\State $t \gets 1$
\While{$t \le T$}
  \State $s_t \gets$ current task stage
  \State $\hat p_t \gets \arg\max_{p\in\mathcal{S}}\Pr(p\mid \mathcal{O}_t,\mathcal{P},s_t)$  \Comment{CoT‐VLM unsafe pattern}
  \State $\alpha \gets$ Scheduler.VLM($\hat p_t,\mathcal{P},s_t$)  \Comment{Cost‐term mask}
  \State sample initial noise $A_t^K \sim \mathcal{N}(0,I)$
  \For{$k = K$ \textbf{down to} $1$}  \Comment{DDIM denoising}
    \State $\mu \gets \mu(A_t^k,\mathcal{O}_t,k)$
    \State $A_{0\mid k} \gets$ Estimate clean chunk via $\epsilon_\theta$
    \If{$k \le M$}  \Comment{apply safety guidance in final $M$ steps}
      \State $\mathcal{C}_{\mathrm{sched}} \gets \sum_i \alpha_i\,\mathcal{C}_i(A_{0\mid k},\mathcal{P},s_t)$
      \State $A_t^{k-1} \gets \mu \;-\;\rho_k\nabla_{A_k}\mathcal{C}_{\mathrm{sched}}\;+\;\sigma_k z,\;z\!\sim\!\mathcal{N}(0,I)$
    \Else
      \State $A_t^{k-1} \gets \mu + \sigma_k z,\;z\!\sim\!\mathcal{N}(0,I)$
    \EndIf
  \EndFor
  \State Execute $A_t^0$ for the next $m$ steps
  \State $t \gets t + m$
\EndWhile
\end{algorithmic}
\end{algorithm}
\vspace{-0.4cm}

% \begin{figure}[!ht]
%   \centering
%   \fbox{%
%     \begin{minipage}[c][5cm][c]{0.97\linewidth}
%       \centering
%       {\Large Dummy figure}
%     \end{minipage}%
%   }
%   \caption{Safety Constrains Guided Denoising Process.}
%   \label{fig:denoising}
% \end{figure}
\section{Implementation Details}
\label{appendix:inference}

\subsection{Real World Setup}
\begin{figure}[!ht]
  \centering
  \vspace{-10pt}
  \includegraphics[width=1.0\linewidth]{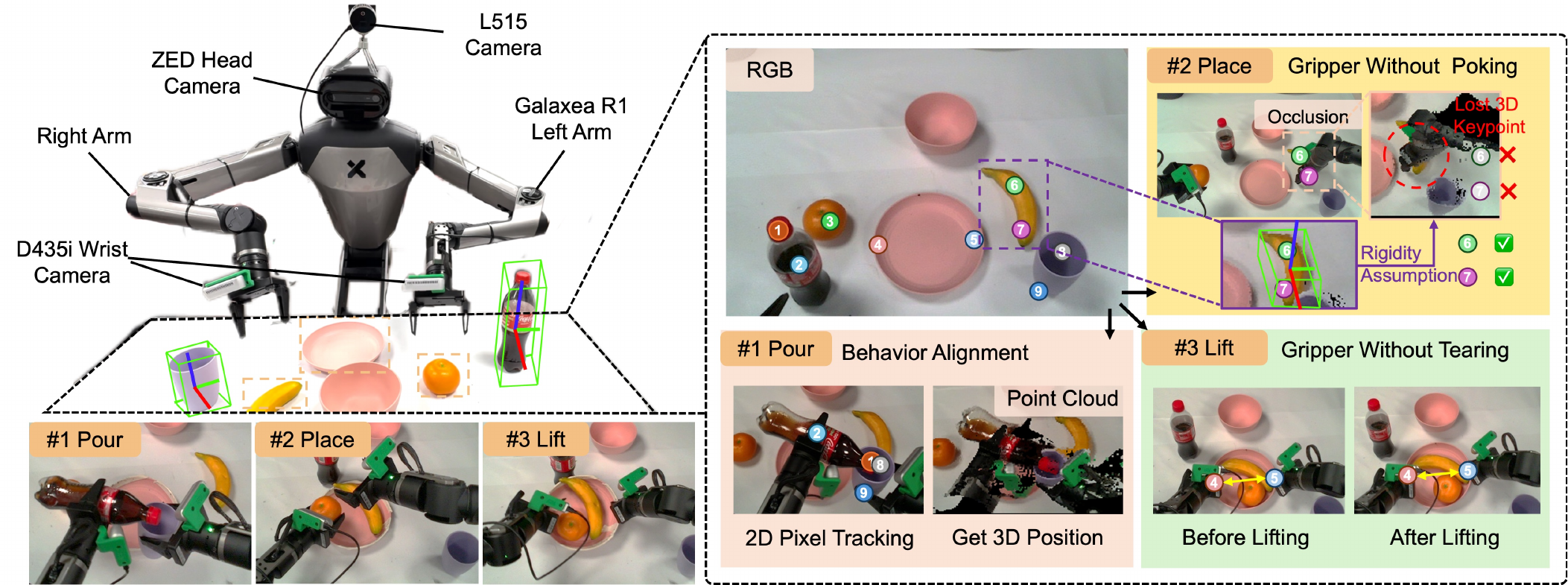}
  \vspace{-10pt}
  \caption{Real-world experimental platforms (Left) and Keypoint Proposal and Tracking (Right) in SafeBimanual.}
  \label{fig:keypoint_proposal}
  \vspace{-0.4cm}
\end{figure}

All experiments use the Galaxea R1 humanoid robot, featuring two 7-DoF arms (70 cm reach, 100 N force), a 4-DoF waist, as shown in Figure~\ref{fig:keypoint_proposal}. Scene rgb perception is provided by a ZED head-camera and wrist-mounted Intel D435i camera, supplemented by an Intel L515 for high-fidelity point clouds. The real-world experiments are performed on a single RTX 4080s GPU.  

\subsection{Training Settings of Baseline Methods}\label{app:params}
The key training setup for our baseline policies is detailed in Table~\ref{tab:baselines_settings}. In simulation experiments, we evaluate 2D Diffusion Policy (DP), 3D Diffusion Policy (DP3), and RDT‑1b; for real‑world tasks, we use DP and improved 3D Diffusion Policy (iDP3) as baselines. All models are trained or fine‑tuned with 50 episodes per task.

\vspace{-0.5cm}
\begin{table}[h]
\centering
\caption{Hyper-parameter Settings for DP, DP3, RDT-1b, and iDP3 Algorithms.}
\label{tab:baselines_settings}
\resizebox{\textwidth}{!}{
\begin{tabular}{lcccc}
\toprule
\textbf{Parameter} & \textbf{DP \cite{chi2023diffusion}} & \textbf{DP3 \cite{ze20243d}} & \textbf{RDT-1b \cite{liu2024rdt}} & \textbf{iDP3 \cite{ze2024humanoid_manipulation}} \\
\midrule
horizon & 8 & 8 & 64 & 16 \\
n.obs\_steps & 3 & 3 & 2 & 2 \\
n.action\_steps & 6 & 6 & 30 & 15 \\
num.inference\_steps & 100 & 10 & 5 & 10 \\
dataloader.batch\_size & 128 & 256 & 32 & 64 \\
dataloader.num\_workers & 0 & 8 & 8 & 8 \\
dataloader.shuffle & True & True & True & True \\
dataloader.pin\_memory & True & True & True & True \\
dataloader.persistent\_workers & False & False & False & False \\
optimizer.target & AdamW & AdamW & AdamW & AdamW \\
optimizer.lr & 1.0e-4 & 1.0e-4 & 1.0e-4 & 1.0e-4 \\
optimizer.betas & [0.95, 0.999] & [0.95, 0.999] & [0.9, 0.999] & [0.95, 0.999] \\
training.lr.scheduler & cosine & cosine & constant & cosine \\
training.lr.warmup\_steps & 500 & 500 & 500 & 500 \\
training.num\_epochs & 300 & 3000 & 10000 & 3000 \\
training.gradient.accumulate.every & 1 & 1 & 1 & 1 \\
training.use.ema & True & True & True & True \\
training.gpu & 6000 Ada & 6000 Ada & A800 & 6000 Ada \\
\bottomrule
\end{tabular}
}
\vspace{-0.2cm}
\end{table}

\subsection{Implementation Details of Keypoint Proposal and Tracking}
\label{app:Keypoints}

\textbf{Keypoint Proposal.}
We refer to these keypoints as \emph{safety constraint primitives}.  
In SafeBimanual, we define safety constraint primitives at both the object and gripper levels to capture all relevant interaction cues (see Figure~\ref{fig:keypoint_proposal}):
\begin{itemize}
  \item \textbf{Object-level keypoint:}  
    Local surface keypoints are extracted from RGB‑D images using DINOv2 features, SAM2 segmentation~\cite{ravi2024sam}, PCA reduction, \(k\)-means clustering (\(k=5\)), and back‑projection into world coordinates (see ReKeP~\cite{huang2024rekep} for more details). Candidates within 4cm of any existing keypoint are then filtered out to ensure spatial diversity. 
    The object center and primary axis come from the translation and principal axis of the 6D pose estimated by Omni6DPose~\cite{zhang2024omni6dpose}, with axis grounding as in OmniManip~\cite{pan2025omnimanip}.
  \item \textbf{Gripper-level keypoint:}  
    Gripper tip positions \(p^{\mathrm{tip}}_i = \mathcal{F}(q^i)\,T_i^{\mathrm{tip}}\) are computed via differentiable forward kinematics, where \(T_i^{\mathrm{tip}}\) is the fixed offset from end‑effector to tip. We do not perform surface segmentation or generate additional local keypoints on the gripper itself, which simplifies computation and analysis while still ensuring effective prevention of gripper collisions.
\end{itemize}

These object-level and gripper-level primitives are used to compute our five differentiable safety costs, enabling modular, interpretable, and dynamically scheduled safety constraints during diffusion‑guided trajectory optimization.

\textbf{Keypoint Tracking Under Occlusion.}  
Although CoTracker3~\cite{karaev24cotracker3} provides robust 2D pixel‐level tracking under heavy occlusions, direct 3D point‐cloud measurements can still be lost. To recover occluded 3D keypoints, we assume each local surface keypoint remains fixed in the object’s frame. Concretely, let \(T_{wo}(t)\in\mathrm{SE}(3)\) be the object’s pose in the world frame at time \(t\) (from Omni6DPose~\cite{zhang2024omni6dpose}), and let \(p_k^o\in\mathbb{R}^3\) be the keypoint coordinates in that object frame. Then its world coordinate is \(k_t = T_{wo}(t)\,p_k^o\). When first observed at \(t_0\), we initialize \(p_k^o = T_{wo}(t_0)^{-1}k_{t_0}\). For any later timestep \(t\) where the point is missing, we reconstruct it as \(\widehat{k}_t = T_{wo}(t)\,p_k^o\). If a noisy observation \(k_t^{\mathrm{obs}}\) becomes available, we optionally refine via exponential smoothing:  
\begin{equation}
p_k^o \leftarrow \beta\,p_k^o + (1-\beta)\,T_{wo}(t)^{-1}k_t^{\mathrm{obs}},\quad \beta\in[0,1].
\label{eq:Occlusion_keypoint}
\end{equation}
This static‐offset assumption ensures accurate, temporally consistent 3D keypoint estimates even under severe occlusions.

\subsection{CoT-based Vision-Language Model Prompts}\label{app:Prompt}
After identifying keypoint candidates, we number them \(\{0, \dots, K{-}1\}\) and overlay them on the key-frame RGB image, which is then input, along with a high-level task instruction, into the Vision-Language Model (VLM). We adopt a code-style prompt template centered around a \texttt{stage\_n\_llm\_output} dictionary and a \texttt{stage\_n\_end()} function skeleton. Only minimal examples are given to illustrate stage structure and constraint formats; all remaining fields (e.g., number of stages, keypoints per stage, termination logic) are inferred autonomously by the VLM through physical priors and visual understanding.

In this work, GPT-4o is used as the default VLM. As newer VLMs emerge, they can be directly substituted to enhance performance without modifying downstream code. Hence, our goal is not prompt engineering per task type, but to establish a sustainable, end-to-end pipeline: as long as key names and function interfaces remain consistent, newer VLMs can iteratively generate stage-wise constraints that conform to the template format and drive robotic execution in increasingly complex tasks. To illustrate the CoT‐VLM scheduling process, we present a concrete prompt instance that aligns with our two‐stage procedure:

\begin{lstlisting}
## Instructions
## Universal VLM-Prompt Template 
# ==============================================================
# Suppose you are helping a bimanual robot to ensure the safety 
# during performing manipulation tasks by output a python dictionary 
# and stage_end function. The manipulation task is given as an image 
# of the environment, overlayed with keypoints marked with their 
# indices, along with coordinates of keypoints. Your job should be 
# focus on avoiding unsafe behaviors between the objects.

# Inputs: - system prompt(you are safety assistant)  
#         - user RGB image(keypoints numbered)  
#         - user np.array of keypoint XYZ coordinates  

### ---- 1. KEYPOINT ALIAS TABLE -----------------------------
KEYPOINTS = {
    -1 : "left_gripper_tip",
    -2 : "right_gripper_tip",
    -3 : "apple_centre",
    -4 : ...
    # Additional ids are defined by the annotated image
}

# ==============================================================

## GENERAL NOTES
# - Keypoints coordinates input, is an np.array of three-dimensional 
#   point coordinates, each row represents the X Y Z coordinates of a 
#   point, the first row represents the point 1, the second represents 
#   the point 2, and so on.  
# - Choose points by rows, which is point 1, 2, 3... And -1 is 
#   left_gripper_tip while -2 is right_gripper_tip
# - Decide which arm is nearer if only one should act.  
# - Typical guidance types: poking / tear / align / gripper collision / 
#   object collision.  
# - stage_end is omitted for final stage unless needed.

## STEP-BY-STEP INSTRUCTION FOR GPT4o
# 1. Determine how many stages are involved in the task. Grasping 
#    must be an independent stage. 
Some examples:
  - "dual_bottle_easy":
    - 2 stages: "grasp bottles", "move and place bottles"
# 2. For **each** stage:  
#   a. Choose guidance type & fill llm_output. (poking, objects-collision, grippers-collision, tear, align)
#       if gripper is going to grasp object:
#           enable poking_guidance
#       if two objects are both grasped and moving:
#           enable collision_guidance (between objects)
#       if object is not successfully grasped:
#           enable collision_guidance (between grippers)
#       if both grippers are holding the same object:
#           enable tear_guidance
#       if two points on the different objects need to be aligned:
#           enable align_guidance

# llm_output example: 
llm_output = { 
	"enable_?_guidance": 
	{ 
		"enable": True/False,
		"enable_left_arm":  
		{ 
		"enable": True/False, 
		"point": "?",  
		}, 
		"enable_right_arm": 
		{ 
		"enable": True/False, 
		"point": "?",  
		} 
	}, 
}

#   b. If the stage needs a finish condition, implement  
def stage_N_end(self): 
	...
	return ...  
#   **Nmote:**
#   - Consider use bool to determin the start and end of the stage, in 
#     case the stage is enable more than once.
#   c. Write one-line comment explaining key-point choices.  
# 3. Preserve ordering: Stage 1,2,3...
# 4. Return only code no extra narration.

# ==============================================================

## Example A "apple handover"
# Stages: 1 grasp apple, 2 handover
### stage_1 "grasp apple"
# left arm nearer to apple (choose apple centre)
llm_output = {
    "enable_poking_guidance":{
        "enable": True,
        "enable_left_arm":  {"enable": False, "point": "-3"},
        "enable_right_arm": {"enable": True,  "point": null}
    }
}
def stage_1_end(self): 
    return self.is_left_gripper_close() and (self.apple_pose[2] > 0.85)

# both grippers hold apple centre; tips are "-1" (L) & "-2" (R)
llm_output_2T = {
    "enable_tear_guidance":{
        "enable": True,
        "enable_left_arm":  {"enable": True, "point": "-1"},
        "enable_right_arm": {"enable": True, "point": "-2"}
    }
}

# both grippers hold apple centre; tips are "-1" (L) & "-2" (R)
llm_output_2F = {
    "enable_gripper_collision_guidance":{
        "enable": True,
        "enable_left_arm":  {"enable": True, "point": "-1"},
        "enable_right_arm": {"enable": True, "point": "-2"}
    }
}

def stage_tear_guidance_start(self):
    return self.is_left_gripper_close() and self.is_right_gripper_close() and (self.apple_pose[2] > 0.85)

def stage_gripper_collision_guidance_start(self):  
    return (self.is_left_gripper_close() and self.left_gripper.pose()[2] > 0.85 and self.apple_pose[2] < 0.8)

def get_current_guidance(self, llm_output_2T, llm_output_2F):
    # Decide which guidance dictionary to return
    if self.stage_tear_guidance_start():
        return llm_output_2T
    elif self.stage_gripper_collision_guidance_start():
        return llm_output_2F
        
## Example B "pour coke into cup"
# Stages: 1 grasp objects, 2 align, 3 pour
### stage_1 "grasp coke and cup"
# left gripper: kp-3 (coke mid), right: kp-5 (cup mid)
llm_output = {
    "enable_poking_guidance":{
        "enable": True,
        "enable_left_arm":  {"enable": True, "point": "3"},
        "enable_right_arm": {"enable": True, "point": "5"}
    }
}
def stage_1_end(self):
    return self.left_gripper_close() and self.right_gripper_close()

### stage_2 "align coke with cup"
# collision points kp-1 (coke neck) & kp-4 (cup rim)
llm_output = {
    "enable_collision_guidance":{
        "enable": True,
        "enable_left_arm":  {"enable": True, "point": "1"},
        "enable_right_arm": {"enable": True, "point": "4"}
    }
}
def stage_2_end(self):
    height = abs(self.object1_pose[2] - self.object2_pose[2])
    return height > 0.10   # coke 10cm above cup

### stage_3 "pour coke"
# align guidance kp-1 & kp-4 (same as above)
llm_output = {
    "enable_align_guidance":{
        "enable": True,
        "enable_left_arm":  {"enable": True, "point": "1"},
        "enable_right_arm": {"enable": True, "point": "4"}
    }
}

# ----- END OF EMPTY TEMPLATE -----

\end{lstlisting}

As an example, Figure~\ref{fig:keypoint_proposal} illustrates a sub-task of pouring water. After prompting, the Vision-Language Model (VLM) analyzes a keyframe image with keypoints mask and correctly infers the high-risk scenario of \textbf{Objects Behavior Misalignment}, activating the cost term \(\mathcal{C}_2\). Based on its interpretation, the VLM identifies and returns two keypoints: one on the top of the bottle and one on the top of the cup. To initiate a safe and effective pouring motion, a vector \(\vec{a}\) is first constructed between the top point of the cup and the central axis point of the cup. This vector is designed to have the minimum angle with the world frame's vertical z-axis. Next, a vector \(\vec{b}\) is constructed between the top of the cup and the top of the bottle. The robot then executes the pouring motion by aligning \(\vec{b}\) with \(\vec{a}\), thereby ensuring the bottle is tilted in a direction that minimizes spillage and avoids collision. This behavior alignment cost ensures that dynamic constraints are grounded in visual observations and can be flexibly scheduled based on scene semantics and geometry.

\section{Additional Results}
\label{appendix:inference}
\subsection{Visualization of the Generated Safe Bimanual Trajectories in RoboTwin}
\begin{figure}[h]
  \centering
  \vspace{-20pt}
  \includegraphics[width=1.0\linewidth]{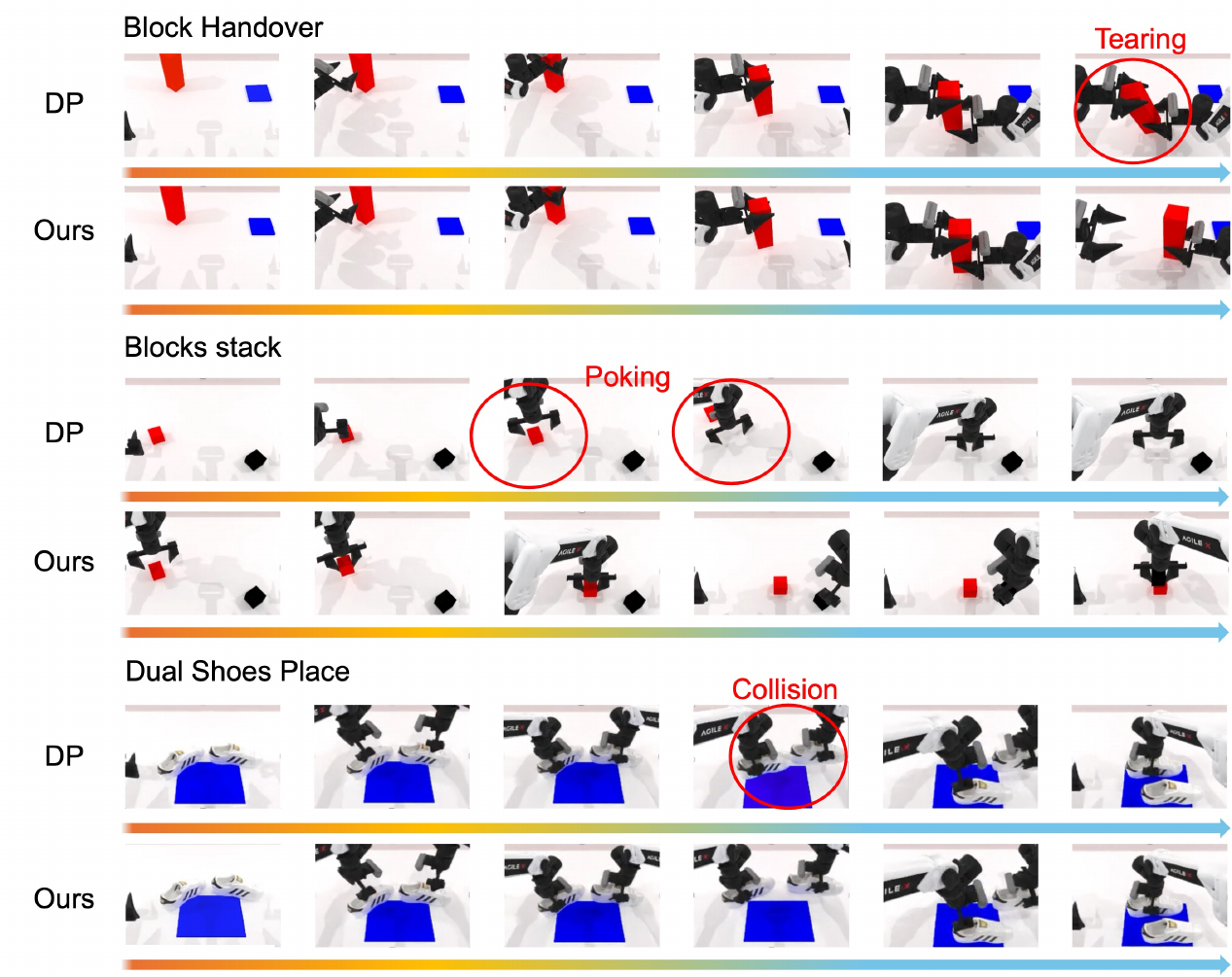}
  \vspace{-10pt}
  \caption{Simulation Tasks Visualization.}
  \label{fig:simulation visualization}
\end{figure}

In Figure~\ref{fig:simulation visualization}, we compare Diffusion Policy outputs against trajectories refined by SafeBimanual across three RoboTwin tasks. \textbf{Block Handover.} During the handover stage, SafeBimanual correctly identifies the risk of gripper tearing and activates \(\mathcal{C}_4\) to maintain a safe inter-tip distance. As a result, the optimized trajectory avoids the tearing seen in the unmodified Diffusion Policy rollout. \textbf{Blocks Stack (Easy).} At the pick stage, SafeBimanual detects potential gripper poking and applies \(\mathcal{C}_3\) to enforce directional alignment between the gripper tip and the cube keypoint. This prevents surface damage and enables a stable grasp. \textbf{Dual Shoes Place.} While lifting the shoes, SafeBimanual recognizes an imminent object–object collision and selects the shoe-head (left) and shoe-heel (right) keypoint pair to compute \(\mathcal{C}_1\). The guided trajectory steers clear of high-risk regions, ensuring both shoes are moved safely to their target areas. 

\vspace{0.4cm}

\subsection{Visualization of the Generated Safe Bimanual Trajectories in Real-world}
\begin{figure}[!ht]
  \centering
  \vspace{-10pt}
  \includegraphics[width=1.0\linewidth]{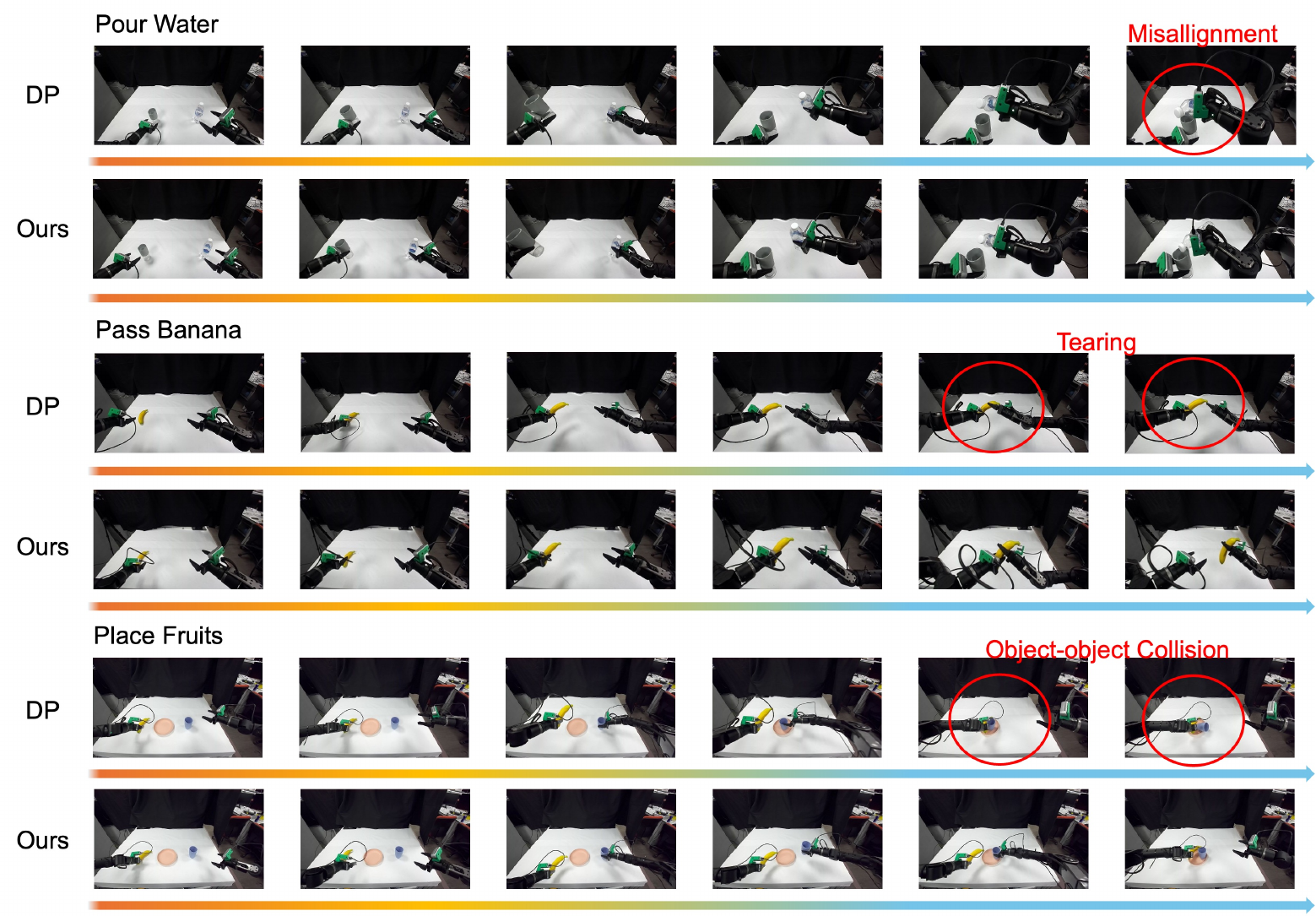}
  \vspace{-10pt}
  \caption{Real-world Tasks Visualization.}
  \label{fig:real-world visualization}
\end{figure}

Figure~\ref{fig:real-world visualization}, we compare Diffusion Policy outputs against trajectories refined by SafeBimanual across three real-world tasks:
\textbf{Pour Water.} During the pouring phase, SafeBimanual infers misalignment and activates \(\mathcal{C}_2\), selecting the bottle mouth and cup mouth keypoints. The guided trajectory corrects the tilt to align spout and cup, preventing spillage.  
\textbf{Pass Banana.} In the handover stage, SafeBimanual detects potential tearing and enables \(\mathcal{C}_4\), enforcing a fixed inter-tip distance as both grippers hold the banana. This avoids relative motion that would damage the fruit.  
\textbf{Place Fruits.} At the final placement phase, SafeBimanual triggers the objects collision cost \(\mathcal{C}_1\), choosing a surface keypoint on the banana and one on the cup. The resulting trajectory maintains safe separation, preventing the banana from knocking over the juice cup.  

% \end{document}

% no \bibliographystyle is required, since the corl style is automatically used.

% \bibliography{reference}% .bib

% \end{document}